%% file: CameraReadyAAAI23.tex
\documentclass[letterpaper]{article} %
\usepackage{aaai23}  %
\usepackage{times}  %
\usepackage{helvet}  %
\usepackage{courier}  %
\usepackage[hyphens]{url}  %
\usepackage{graphicx} %
\usepackage{natbib}  %
\usepackage{caption} %
\usepackage{newfloat}
\usepackage{listings}
\DeclareCaptionStyle{ruled}{labelfont=normalfont,labelsep=colon,strut=off} %
\input{src/00_Preamble}

\title{Unfooling Perturbation-Based Post Hoc Explainers}
\author{
    Zachariah Carmichael, Walter J. Scheirer
}
\affiliations{
    University of Notre Dame\\

    zcarmich@nd.edu, walter.scheirer@nd.edu
}

\begin{document}

\maketitle

\begin{abstract}
  \input{src/01_Abstract}
\end{abstract}

\input{src/02_Introduction}
\input{src/03_Background}
\input{src/04_Methods}
\input{src/05_Experiments}
\input{src/06_Discussion}

\section*{Acknowledgments}
\input{src/07_Acknowledgements}

\bigskip

\bibliography{aaai23}

\appendix
\clearpage
\begin{center}
    {\LARGE Supplemental Material}
\end{center}
\input{src/99_Supplemental}

\end{document}

%% file: src/00_Preamble.tex
\usepackage{booktabs}
\usepackage{multirow}
\usepackage{rotating}
\usepackage{amsmath}
\usepackage{amssymb}
\usepackage{mathtools}
\usepackage{stmaryrd}
\usepackage{bm}
\usepackage{mathrsfs}
\usepackage{enumitem}
\usepackage[
  ruled,
  vlined,
  linesnumbered,
]{algorithm2e}

\usepackage{subcaption}

\usepackage{makecell}

\usepackage{minitoc}

\usepackage{pifont}%
\newcommand{\cmark}{\ding{51}}%
\newcommand{\xmark}{\ding{55}}%

\newcommand{\ra}[1]{\renewcommand{\arraystretch}{#1}}

\DeclareMathOperator*{\argmin}{argmin}
\newcommand{\etal}{et al.}
\newcommand{\ie}{i.e.}
\newcommand{\eg}{e.g.}
\newcommand{\Ie}{I.e.}

\newcommand{\etc}{etc.}
\newcommand{\posthoc}{post hoc}
\newcommand{\PostHoc}{Post Hoc}
\newcommand{\Posthoc}{Post hoc}
\newcommand{\SHAP}{{\small\texttt{SHAP}}}
\newcommand{\NeighborhoodSHAP}{{\small\texttt{Neighborhood\,SHAP}}}
\newcommand{\LIME}{{\small\texttt{LIME}}}
\newcommand{\CLIME}{{\small\texttt{CLIME}}}
\newcommand{\ACME}{ACME}
\newcommand{\dood}{d}  %
\newcommand{\ginline}{\text{\tiny(}g\text{\tiny)}}
\newcommand{\hinline}{\text{\tiny(}h\text{\tiny)}}
\newcommand{\KNNCAD}{{\small\texttt{KNN-CAD}}}
\newcommand{\DETECT}{{\small\texttt{CAD-Detect}}}
\newcommand{\DETECTtiny}{\shortstack{\texttt{\tiny{CAD-}}\\\texttt{\tiny{Detect}}}}
\newcommand{\DEFENSE}{{\small\texttt{CAD-Defend}}}
\newcommand{\DEFENSEtiny}{\shortstack{\texttt{\tiny{CAD-}}\\\texttt{\tiny{Defend}}}}

\let\oldeqref\eqref
\renewcommand{\eqref}[1]{Eq.\ \oldeqref{#1}}

\usepackage[breaklinks=true,colorlinks,bookmarks=false]{hyperref}
\usepackage{float}

%% file: src/01_Abstract.tex
Monumental advancements in artificial intelligence~(AI) have lured the interest of doctors, lenders, judges, and other professionals. While these high-stakes decision-makers are optimistic about the technology, those familiar with AI systems are wary about the lack of transparency of its decision-making processes.
Perturbation-based \posthoc{} explainers offer a model agnostic means of interpreting these systems while only requiring query-level access. However, recent work demonstrates that these explainers can be fooled adversarially. This discovery has adverse implications for auditors, regulators, and other sentinels.
With this in mind, several natural questions arise -- how can we audit these black box systems? And how can we ascertain that the auditee is complying with the audit in good faith?
In this work, we rigorously formalize this problem and devise a defense against adversarial attacks on perturbation-based explainers.
We propose algorithms for the detection (\DETECT{}) and defense (\DEFENSE{}) of these attacks, which are aided by our novel conditional anomaly detection approach, \KNNCAD{}.
We demonstrate that our approach successfully detects whether a black box system adversarially conceals its decision-making process and mitigates the adversarial attack on real-world data for the prevalent explainers, \LIME{} and \SHAP{}.\footnote{Code for this work is available at \url{https://github.com/craymichael/unfooling}}

%% file: src/02_Introduction.tex
\section{Introduction}

As a result of the many recent advancements in artificial intelligence~(AI), a significant interest in the technology has developed from high-stakes decision-makers in industries such as medicine, finance, and the legal system~\cite{liptonMythos2018,millerAIMedical2018,Rudin2019}.
However, many modern AI systems are black boxes, obscuring undesirable biases and hiding their deficiencies~\cite{adversarialExamples,hendrycksNaturalAdversarialExamples2019,liptonMythos2018}.
This has resulted in unexpected consequences when these systems are deployed in the real world~\cite{oneilWeaponsMathDestruction2016,buolamwiniGenderShadesIntersectional2018,incidentDB}.
Accordingly, regulatory and legal ordinance has been proposed and implemented~\cite{EU-GDPR,EU-US-TTC_statement2021,EU-AI-Act}.

All of this naturally leads to the question: how can we audit opaque algorithms?
\Posthoc{} explanation methods offer a way of understanding black box decision-making processes by estimating the influence of each variable on the decision value.
Unfortunately, there is a medley of incentives that may motivate an organization to withhold this information, whether it is financial, political, personal, or otherwise.
Indeed, these explainers are demonstrably deceivable~\cite{fooling,adv_xai_site} --- if an organization is aware that its algorithms are under scrutiny, it is capable of falsifying how its algorithms operate.
This begs the question --- how do we know that the auditee is complying faithfully?

In this work, we explore the problem of employing perturbation-based \posthoc{} explainers to audit black box algorithms.
To the best of our knowledge, this is the first paper to address the aforementioned questions and provide a solution.
We explore a multi-faceted problem setting in which the auditor needs to ascertain that the algorithms of an organization: 1) do not violate regulations or laws in their decision-making processes and 2) do not adversarially mask said processes.
Our contributions are as follows:
\begin{itemize}[noitemsep,nolistsep,leftmargin=1em]
    \item We formalize real-world adversarial attack and defense models for the auditing of black box algorithms with perturbation-based \posthoc{} explainers. We then formalize the general defense problem against adversarial attacks on explainers, as well as against the pragmatic scaffolding-based adversarial attack on explainers~\cite{fooling}.
    \item We propose a novel unsupervised conditional anomaly detection algorithm based on \textit{k}-nearest neighbors: \KNNCAD{}.
    \item We propose adversarial detection and defense algorithms for perturbation-based explainers based on any conditional anomaly detector: \DETECT{} and \DEFENSE{}, respectively.
    \item Our approach is evaluated on several high-stakes real-world data sets for the popular perturbation-based explainers, \LIME{}~\cite{lime} and \SHAP{}~\cite{shap}. We demonstrate that the detection and defense approaches using \KNNCAD{} are capable of detecting whether black box models are adversarially concealing the features used in their decision-making processes. Furthermore, we show that our method exposes the features that the adversaries attempted to mask, mitigating the scaffolding-based attack.
    \item We introduce several new metrics to evaluate the fidelity of attacks and defenses with respect to explanations and the black box model.
    \item We conduct analyses of the explanation fidelity, hyperparameters, and sample-efficiency of our approach.
\end{itemize}

%% file: src/03_Background.tex
\section{Background}\label{sec:background}

\paragraph{Local Black Box \PostHoc{} Explainers}
Of particular relevance to auditing an algorithm is in understanding individual decisions.
Explainable AI~(XAI) approaches afford transparency of otherwise uninterpretable algorithms~\cite{BARREDOARRIETA202082} and are conducive to auditing~\cite{Akpinar2022,ZHANG2022100572}.
Local \posthoc{} explainers, notably \LIME{} and \SHAP{}, do so by estimating the contribution of each feature to a decision value. These particular explainers are classified as model agnostic and black box, which satisfies the conditions of an audit --- the auditor has no \textit{a priori} knowledge of the model class and has only query-level access.
Both approaches produce explanations by fitting linear models to a dataset generated by perturbing the neighborhood about a sample.
Let $\bm{\mathcal{D}} = (\bm{\mathcal{X}} \times \bm{\mathcal{Y}}) = \{(\mathbf{x}_1, {y}_1), (\mathbf{x}_2, {y}_2), \dots, (\mathbf{x}_N, {y}_N)\}$ be the data set
where each data sample $\mathbf{x}_i \in \mathbb{R}^F$ has $F$ features
and each label ${y}_i \in \mathbb{N}_{0}$ represents one of $C$ classes encoded as an integer in the range $[0..C)$.
We denote the black box classifier as $f: \bm{\mathcal{X}} \rightarrow \bm{\mathcal{Y}}$ and the explainer as $g$.
The general problem these explainers solve in order to explain an instance $\mathbf{x}_i$ is given by \eqref{eq:limeshap_problem}
\begin{equation}\label{eq:limeshap_problem}
    \argmin_{g_{\mathbf{x}_i} \in \mathcal{G}}
    \hspace{-2pt}
    \sum_{\mathbf{x}^{(g)}_j \in \bm{\mathcal{X}}_i^{(g)}}
    \hspace{-4pt}
    \left( f(\mathbf{x}^{(g)}_j) - g_{\mathbf{x}_i}(\mathbf{x}^{(g)}_j) \right) ^ 2
    \hspace{-2pt}
    \pi_{\mathbf{x}_i}(\mathbf{x}^{(g)}_j) \,+\, \Omega(g_{\mathbf{x}_i})
\end{equation}
\noindent
The minimization objective is a function of the linear model $g_{\mathbf{x}_i}$ from the set of all linear models $\mathcal{G}$, the neighborhood function $\pi_{\mathbf{x}_i}$, and the regularization function  $\Omega$.
With a slight abuse of notation, $\pi_{\mathbf{x}_i}$ both generates the neighborhood of $\mathbf{x}_i$ ($\bm{\mathcal{X}}_i^{\ginline{}}$) and gives the proximity of each $\mathbf{x}^{\ginline{}}_j$ to $\mathbf{x}_i$. The latter two functions are defined using game-theoretic means for \SHAP{} and empirical means for \LIME{}.
Both explainers produce explanations as a set of feature contributions $\mathcal{E}_i=\{a_{ij}\}_{j=1}^F$ that describes the contribution, or importance, of each feature to the decision value $y_i$.
In our notation, $a_{ij}$ indicates the explained contribution of the $j^{\text{th}}$ feature to $y_i$.

\paragraph{Adversarial Attacks on Explainers}

Szegedy \etal{}\ were the first to discover the phenomena of \textit{adversarial examples} \cite{adversarialExamples} --- human-indistinguishable manipulation of the input space that results in misclassification.
It is worth noting that naturally-occurring adversarial examples have been shown to exist~\cite{hendrycksNaturalAdversarialExamples2019}, though the focus of this work concerns human adversaries.
In contrast to the literature on neural network adversarial attacks~\cite{adversarialattackdefense2020}, attacks on explainers are comparatively less explored, especially those that are perturbation-based~\cite{adv_xai_site}.
Slack \etal{}\ were the first to propose an adversarial attack on black box \posthoc{} explainers~\cite{fooling}. The motivation for the attack arose from the observation that explainer perturbations are often off of the data manifold. In the attack, the adversary aims to detect whether some $\mathbf{x}_i$ originates from $\bm{\mathcal{X}}$ or from the explainer-generated samples, $\bm{\mathcal{X}}^{\ginline{}}$, using a scaffolding classifier.
$\bm{\mathcal{X}}^{\ginline{}}$ is generated by perturbing each $\mathbf{x}_i \in \bm{\mathcal{X}}$, creating a set of new instances $\bm{\mathcal{X}}_i^{\ginline{}}$. The union of each $\bm{\mathcal{X}}_i^{\ginline{}}$ yields the complete set, \ie{}, $\bm{\mathcal{X}}^{\ginline{}} = \bigcup_i \bm{\mathcal{X}}^{\ginline{}}_i$.
If the adversary determines that the query originates from an explainer, then it can elect to predict using an \textit{unbiased} model. In effect, the adversary lies to the explainer about how it uses the features.
Formally, given a biased predictor $f_{\text{biased}}$ and an unbiased predictor $f_{\text{unbiased}}$, the adversary can toggle between predictors using an out-of-distribution~(ood) detector $\dood{}$ according to~\eqref{eq:adv_scaffold}.
\begin{equation}\label{eq:adv_scaffold}
    f(\mathbf{x}_i) =
    \begin{cases}
        f_{\text{biased}}(\mathbf{x}_i)    & \text{if }\dood{}(\mathbf{x}_i) = 1 \\
        f_{\text{unbiased}}(\mathbf{x}_i) & \text{otherwise}
    \end{cases}
\end{equation}
The detector $\dood{}$ is trained on the samples $\bm{\mathcal{X}} \cup \bm{\mathcal{X}}^{\ginline{}}$ with a corresponding label of `0' if the sample is ood (from $\bm{\mathcal{X}}^{\ginline{}}$) and `1' otherwise (from $\bm{\mathcal{X}}$).
Put simply, the detector $\dood{}$ determines if the input sample originated from real data or is an explainer-generated perturbation.
If the sample origin is predicted to be an explainer, the unbiased predictor $f_{\text{unbiased}}$ makes predictions in an innocuous manner, \eg{}, independent of sensitive features such as race. The authors demonstrate that the attack obscures the biased use of features, such as making predictions based on race, from \LIME{} and \SHAP{} while maintaining near-perfect prediction fidelity on in-distribution samples.

In related but tangential works, attacks have been developed on the model and explainer simultaneously.
Abdukhamidov \etal{}\ introduce a gradient-free black box attack on XAI systems that manipulates model predictions without significantly altering the explanations by \posthoc{} explainers~\cite{abdukhamidov2022black}.
Closely related, Zhan \etal{}\ develop a joint attack with the same implications~\cite{zhan2022}.
Both works consider explainers that require gradient-level access to the model and are unsuitable for auditing.
Noppel \etal{}\ propose a trigger-based neural backdoor attack on XAI systems that simultaneously manipulates the prediction and explanation of
gradient-based
explainers~\cite{noppelneuralbackdoor2022}. Again, the attack scenario in our work deviates from this model.

\paragraph{Adversarial Defense for Explainers}

In a similar fashion, defense against adversarial attacks is well explored in the literature~\cite{adversarialattackdefense2020}. However, there is relatively scarce work in defending against adversarial attacks on explainers.
Ghalebikesabi \etal{}\ address the problems with the locality of generated samples by perturbation-based \posthoc{} explainers~\cite{ghalebikesabi2021locality}. They propose two variants of \SHAP{}, the most relevant being \NeighborhoodSHAP{} which considers local reference populations to improve on-manifold sampling.
Their approach is able to
mitigate the scaffolding attack~\cite{fooling}. While this is a notable achievement, it is unclear how the approach compares to baseline \SHAP{} with respect to quantitative and qualitative measures of explanation quality, whether it still upholds the properties of \SHAP{} explanations, and other concerns~\cite{ghalebikesabi2021on}.
Also related is a constraint-driven variant of \LIME{}, \CLIME{}~\cite{CLIME}.
\CLIME{} has been demonstrated to mitigate the scaffolding attack, but requires hand-crafted constraints based on
domain expert knowledge and for data to be discrete.

A step forward in explainer defense, Schneider~\etal{}\ propose two approaches to detect manipulated Grad-CAM explanations~\cite{deceptiveAIExpl2022}. The first is a supervised approach that determines if there is (in)consistency between the explanations of an ensemble of models
and explanations that are labeled as manipulated or not. The second is an unsupervised approach that determines if the explanations of $f$ are as sufficient to reproduce its predictions as those from an ensemble of models.
They conclude that detection without domain knowledge is difficult.
Aside from the deviant attack model in their work, we do not require that deceptive explanations be labeled (an expensive and error-prone process) and we require that only a single model be learned (the authors use 35 CNNs as the ensemble in experiments).

Recently, a perturbation-based explainer coined {\small \texttt{EMaP}} was introduced that also helps to mitigate the scaffolding attack~\cite{emap}. Similar to \NeighborhoodSHAP{}, it improves upon its perturbation strategy to create more in-distribution samples. This is accomplished by perturbing along orthogonal directions of the input manifold, which is demonstrated to maintain the data topology more faithfully.

\paragraph{Conditional Anomaly Detection}
Vanilla anomaly detection aims to discover observations that deviate from a notion of normality, typically in an unsupervised paradigm~\cite{anomalyDetectSurvey2021}. Of interest in this work is conditional, also referred to as contextual, anomaly detection. This type of anomaly is an observation that is abnormal in a particular context, \eg{}, in time or space.
Formally, a set of conditional anomalies $\mathcal{A}$ is given by \eqref{eq:conditional_anomalies}
\begin{equation}\label{eq:conditional_anomalies}
    \mathcal{A} = \{ (\mathbf{x}_i, \mathbf{y}_i) \in (\bm{\mathcal{X}} \times \bm{\mathcal{Y}}) \mid \mathbb{P}(\mathbf{y}_i \mid \mathbf{x}_i) \leq \tau \}, ~ \tau \geq 0
\end{equation}
where $\mathbb{P}$ is the probability measure of some probability density function~(pdf) that characterizes
normality and $\tau$ is a low-probability threshold separating normal and abnormal observations.
Following the terminology in \cite{cad}, $\bm{\mathcal{X}}$ is the set of environmental variables that the set of observed variables $\bm{\mathcal{Y}}$ is conditioned on.

Song \etal{}\ proposed the first conditional anomaly detector using Gaussian mixture models --- two sets of Gaussians model the environmental and observed variables, respectively, while a learned probability mapping function determines how the Gaussians in each set map to one another~\cite{cad}.
Since, several approaches have been proposed based on classical and deep learning techniques --- we point to this comprehensive survey for further reading~\cite{anomalyDetectSurvey2021}.
In this work, we propose a new conditional anomaly detection method as 1) the deep learning techniques are data-hungry
and 2) most techniques do not consider or fair well with categorical data, which is plentiful in real-world high-stakes data: credit scoring~\cite{ficoheloc2018}, recidivism risk scoring~\cite{angwin2016machine}, \etc{}

%% file: src/04_Methods.tex
\section{The Problem and a Solution}\label{sec:methods}

\begin{figure}
    \centering
    \includegraphics[width=\linewidth]{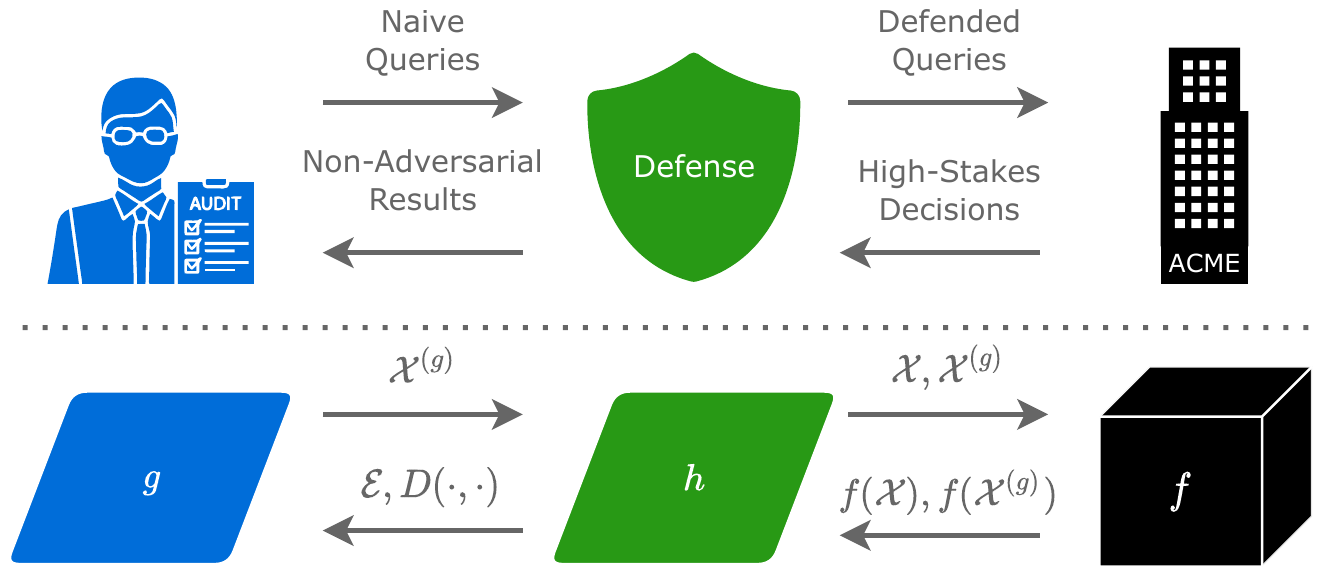}
    \caption{
    An overview of the adversarial attack and defense scenario.
    The top panel mirrors the bottom panel formalism on a higher level. Our defense approach provides defense to auditors from potential adversarial behavior of the auditee, \ACME{}. Given the explainer $g$-generated samples $\bm{\mathcal{X}}^{\ginline{}}$ and reference samples $\bm{\mathcal{X}}$, the defense $h$ queries the black box $f$. With the results, $h$ gives the auditor the defended explanations $\mathcal{E}$ and a measure of adversarial behavior $D(\cdot, \cdot)$.}
    \label{fig:overview}
\end{figure}

\paragraph{The Attack and Defense Models}

In the attack scenario, an organization algorithmically makes decisions on behalf of users (customers, patients, employees, \etc{}) using a black box predictor $f$. Note that $f$ can be a black box due to it being either uninterpretable or proprietary (or both)~\cite{Rudin2019}.
We will refer to the organization as the \textit{\ACME{}} (\textit{Corporation}) for the remainder of the paper.
We recommend referring back to Figure~\ref{fig:overview} as a guide on how elements of the scenario relate.
The attacker only provides query-level access to $f$, \eg{}, via an API, a web form, or even a social media app facial filter. In the case of $f$ being a classifier, only the final decision value is available, not the class-wise probabilities. Moreover, the class of model that $f$ belongs to is assumed to be unknown.
Thus, to ensure that \ACME{} meets legal or regulatory compliance, an auditor uses a local \posthoc{} explainer, $g$, to determine how features are used to make individual decisions.
Due to a variety of incentives (financial, political, personal, \etc{}), an adversary within \ACME{} desires to conceal the behavior of $f$.
The adversary does not know what explainer is employed but may exploit weaknesses that are shared among common black box \posthoc{} explanation algorithms.
Furthermore, the adversary is unaware of when the audit occurs and is only aware of the information contained within the queries made to $f$.
The general attack objective is to minimize the multi-objective problem given by~\eqref{eq:attack_objective_general}
\begin{align}\label{eq:attack_objective_general}
\begin{split}
    \min_{\mathbf{x}_i \in \bm{\mathcal{X}}^{(g)} \cup \bm{\mathcal{X}}}
        \hspace{-2.2pt}\Big(&
            L_{f}\left(
                f(\mathbf{x}_i),
                y_i
            \right),
            |\{
                j \mid a_{ij} {\in} \mathcal{E}_i^{\text{sensitive}},\\&
                r(\mathcal{E}_i^{-}, a_{ij}) {<} r(\mathcal{E}_i^{-}, a_{ik}), \forall a_{ik} {\in} \mathcal{E}_i^{\text{harmless}}
            \}|,
            \\& \max(\mathcal{E}_i^{\text{sensitive}})
        \Big)
\end{split}
\end{align}
\noindent
where $\mathcal{E}_i^{-} {=} \mathcal{E}_i^{\text{sensitive}} \cup \mathcal{E}_i^{\text{harmless}} {=} g(\mathbf{x}_i; f)$ is an ordered set of feature contributions for the $i^\text{th}$ sample, $\mathcal{E}_i^{\text{sensitive}}$ and $\mathcal{E}_i^{\text{harmless}}$ are the contributions for the sensitive and harmless features, respectively, $L_f$ is some metric that measures the error
between its two scalar arguments, and $r$ gives the index of its second argument within its first argument. $\mathcal{E}_i^{-}$ is ordered by decreasing value of the contribution magnitude to the negative (adverse) outcome. Put simply, the first objective quantifies the error between the predictor and the ground truth, the second objective is the number of feature contributions for sensitive features that are greater than those of harmless features, and the third objective quantifies the magnitude of the feature contributions for sensitive features.
The advantage of this formulation is that it applies to any type of attack that intends to manipulate the attribution of features during an audit. \Ie{}, it is not tied to just the scaffolding attack.

In this work, the scaffolding attack is employed by \ACME{} as described in Section~\ref{sec:background}.
This attack is multi-objective and aims to minimize \eqref{eq:attack_objective_general} by finding some $\dood{} \in \mathcal{D}_{\text{ood}}$ that minimizes~\eqref{eq:attack_objective}
\begin{align}\label{eq:attack_objective}
\begin{split}
    \hspace{-5pt}
    \min_{(\mathbf{x}_i, t_i) \in (\bm{\mathcal{X}}^{(g)} \cup \bm{\mathcal{X}}, \mathcal{T})}
        \hspace{-3pt}\Big(\hspace{-1pt}&
            \llbracket t_i{=}0 \rrbracket
                L_{f}\hspace{-2pt}\left(
                    f_{\text{biased}}(\mathbf{x}_i),
                    f(\mathbf{x}_i)
                \right),\\&
            \llbracket t_i{=}1 \rrbracket
                L_{a}\hspace{-2pt}\left(
                    g(\mathbf{x}_i; f_{\text{unbiased}}),
                    g(\mathbf{x}_i; f)
                \right)
        \hspace{-3pt}\Big)
\end{split}
\end{align}
\noindent
where $\llbracket\cdot\rrbracket$ are Iverson brackets, $\mathcal{D}_{\text{ood}}$ is the set of all out-of-distribution detectors, $\mathcal{T}$ is a set of flags indicating whether $\mathbf{x}_i \in \bm{\mathcal{X}}^{\ginline{}}$, and $L_a$ is some metric that measures the error between its two scalar arguments.
Naturally, there is a trade-off between prediction fidelity (objective one) and unbiased explanation to an auditor (objective two).
Section~\ref{sec:experiments} describes how $f_{\text{biased}}$ and $f_{\text{unbiased}}$ are selected.

We now formulate the detection and defense objectives --- here, the goal of \textit{detection} is to determine whether $f$ is using the scaffolding attack, whereas the goal of \textit{defense} is to ascertain whether the scaffolding attack is used for individual samples. As we will detail in the next section, the latter can be used to ``unfool'' the explainers.
Formally, the general sample-wise defense objective is given by \eqref{eq:defense_problem}
\begin{equation}\label{eq:defense_problem}
    \argmin_{h \in \mathcal{H}} L_{h}\left( h(\mathbf{x}_i; f), \mathbb{P}\left( f(\mathbf{x}_i) \mid \mathbf{x}_i \right) \right) \text{ s.t. }\mathbf{x}_i \in \bm{\mathcal{X}}^{(g)} {\cup} \bm{\mathcal{X}}
\end{equation}
\noindent
where $\mathcal{H}$ is the class of conditional distribution estimators, $h$ is the detector that estimates the conditional pmf $p$ with support $\{0, 1\}$,
and $\mathbb{P}$ is a probability measure of $p$\footnote{To conserve space, we omit random variables of arguments to the pmf $p$ (and $\mathbb{P}$) and use realizations instead.}. The pmf $p$ is the distribution of normality for $f$.
Here, $L_{h}$ is some metric that measures the error between its two scalar arguments.
The role of $h$ is to determine the likelihood of a prediction by $f$ given a sample.
The threshold-based detection is in turn used to classify anomalies for individual samples as in~\eqref{eq:conditional_anomalies}.

To globally detect an attack, the same objective should be minimized as for defense. However, discriminating an attack instead is determined by having $h$ estimate~\eqref{eq:detection_classification}
\begin{equation}\label{eq:detection_classification}
    \left\llbracket D\left(
        p( f(\mathbf{x}_i) \mid \mathbf{x}_i ),
        p( f(\mathbf{x}^{(g)}_i) \mid \mathbf{x}^{(g)}_i )
    \right) \geq \tau_{\text{global}} \right\rrbracket %
\end{equation}
\noindent
where $\tau_{\text{global}} \geq 0$ is a threshold, $\mathbf{x}_i^{\ginline{}} {\in} \bm{\mathcal{X}}^{\ginline{}}$, $\mathbf{x}_i {\in} \bm{\mathcal{X}}$, and $D$ is an asymmetric measure of statistical distance between the two distributions that is permitted to take on negative values and maintains the identity of indiscernibles (\ie{}, $D(x, y) {=} 0 \text{ iff } x {=} y$). With a properly calibrated $h$ and sufficient samples to represent the
distribution of normality,
$D(\cdot, \cdot) > 0$ if $\dood{}(\mathbf{x}_i){=}1$ (or if $f$ is not adversarial) and $D(\cdot, \cdot) \leq 0$ otherwise.

\paragraph{Detection, Defense and \KNNCAD{}}

In this section, we describe a non-parametric approach to detect conditional anomalies based on \textit{k}-nearest neighbors: \underline{\textit{k}}-\underline{n}earest \underline{n}eighbors \underline{c}onditional \underline{a}nomaly \underline{d}etector (\KNNCAD{}).
We then describe general algorithms for the detection (\DETECT{}) and defense (\DEFENSE{}) of adversarial attacks on $g$ for any given $h$.
For simplicity, we treat $f$ as a classifier in describing \KNNCAD{} --- in the case that $f$ is a regressor, we cast it to a classifier by binning its output.
Since $f$ does not return class-wise probabilities in the problem setting, we exploit the fact that we have a single discrete observed variable $y_i$.
On a high level, the main idea of \KNNCAD{} is to compare the labels of the neighbors of some $\mathbf{x}_i$ to $f(\mathbf{x}_i)$ --- the disagreement of the labels of its neighbors determines the degree of abnormality.
Algorithms~\ref{alg:knncad_fit} (\KNNCAD{}\texttt{.fit}) and~\ref{alg:knncad_score} (\KNNCAD{}{\small\texttt{.score\_samples}}) formalize this process.
With samples representing normality $\bm{\mathcal{X}}$, the standard \textit{k}-nearest neighbors algorithm is fit to the data in \KNNCAD{}\texttt{.fit}. These samples are collected by the auditor and are very unlikely to overlap with data that $f$ (and $\dood{}$, if applicable) were trained on.
After the fit is made, each $\mathbf{x}_i \in \bm{\mathcal{X}}$ is scored by \KNNCAD{}{\small\texttt{.score\_samples}}, which gives the scored samples $\mathcal{S}$ (lower is more abnormal).
Subsequently, the threshold $\tau$ is set to the percentile $\epsilon$ of $\mathcal{S}$. The rounding operator used in
computing the percentile is denoted as
$\text{round}({\cdot})$.

In \KNNCAD{}{\small\texttt{.score\_samples}} (Algorithm~\ref{alg:knncad_score}), the pmf $p(f(\mathbf{x}_i) \mid \mathbf{x}_i)$ is estimated.
To do so, the $k$-nearest neighbors of, and distances from, each $\mathbf{x}_i$ are computed.
For each $\mathbf{x}_i$, the labels of its neighbors are retrieved. For the neighbors belonging to each class, the corresponding distances are gathered (denoted by, \eg{}, $\mathbf{d}_0$ for each $y_i=0$) and then aggregated by the function $\phi$. The aggregator $\phi$ estimates the statistical distance between $\mathbf{x}_i$ and its neighbors by computing, \eg{}, the median, mean, or maximum value.
If the vector argument of $\phi$ is empty, then the output of the function is $\infty$.
In the final step, we are interested in measuring the dynamic range between the aggregate distances corresponding to the label of the queried sample, $d_{y_j}$, and to the alternative label(s), $d_{\neg{}y_j}$. The dynamic range is defined as the ratio between two values on the logarithmic scale as in~\eqref{eq:dynamic_range}.
\begin{equation}\label{eq:dynamic_range}
    \texttt{dynamic\_range}(a, b) = \log\left(\frac{a}{b}\right)
\end{equation}
The values given by the dynamic range of such distances are treated as logits, so we apply the standard logistic function $\sigma$ to map the values to probabilities as in \eqref{eq:logits_to_probs}
\begin{align}\label{eq:logits_to_probs}
\begin{split}
    \zeta(d_{\neg{}y_j}, d_{y_j})
        &= \sigma(\texttt{dynamic\_range}(d_{\neg{}y_j}, d_{y_j})) \\
        &= \frac{1}{1 + \exp(-\log\left(\frac{d_{\neg{}y_j}}{d_{y_j}}\right))} \\ 
        &= \frac{1}{1 + \frac{d_{y_j}}{d_{\neg{}y_j}}} = \frac{d_{\neg{}y_j}}{d_{\neg{}y_j} + d_{y_j}}.
\end{split}
\end{align}
We denote this probability mapping function as $\zeta$.
The algorithm \KNNCAD{}{\small\texttt{.score\_samples}} is written for the case of $f$ being a binary classifier, but it is easily generalizable to multi-class by taking the expected value of the comparison of $d_{y_j}$ to every other ${d}_{\neg{}y_j}$, \ie{}, $\mathbb{E}_{d_{\neg{}y_j} \in \mathbf{d}_{\neg{}y_j}}[\zeta(d_{\neg{}y_j}, d_{y_j})]$.

\input{algs/knncad}

In Algorithm~\ref{alg:adv_detect}, the procedure for adversarial attack detection, \DETECT{}, is detailed.
First, the samples $\bm{\mathcal{X}}$ representing normality are partitioned into $\bm{\mathcal{X}}_{\text{train}}$ ($n_{\text{train}}$ samples) and $\bm{\mathcal{X}}_{\text{test}}$ ($|\bm{\mathcal{X}}| {-} n_{\text{train}}$ samples).
Thereafter, $h$ is fit using $\bm{\mathcal{X}}_{\text{train}}$ and the (hyper-)parameters pertinent to $h$, $\Theta_h$.
After $\bm{\mathcal{X}}^{\ginline{}}_{\text{test}}$ is collected, $h$ estimates
$p(f(\mathbf{x}_i) \mid \mathbf{x}_i)$ and $p(f(\mathbf{x}^{\ginline{}}_i) \mid \mathbf{x}^{\ginline{}}_i)$ for $\bm{\mathcal{X}}_{\text{test}}$ and $\bm{\mathcal{X}}^{\ginline{}}_{\text{test}}$, respectively.
The statistical distance $D$ between these two conditional distributions is estimated by comparing the areas under the empirical cumulative distributions of $\mathcal{S}_{\text{test}}$ and $\mathcal{S}_{\text{test}}^{\ginline{}}$. We denote these areas as $a_{\text{test}}$ and $a_{\text{test}}^{\ginline{}}$, respectively.
In the algorithm, we denote the empirical cumulative distribution function as \texttt{ECDF} and the function that computes the area under a curve using the trapezoidal rule as \texttt{AUC}.
Intuitively, a greater difference between $a_{\text{test}}^{\ginline{}}$ and $a_{\text{test}}$ indicates a concentrated region of low-probability samples (likely adversary behavior of $f$), whereas a smaller difference indicates that an attack was likely not carried out.
In practice, the number of perturbed samples generated around each sample, $n_{p}$, is relatively large --- fewer samples are
necessary to accurately estimate $p$. We explore the sample-efficiency of $h$ in Section~\ref{sec:experiments}.

The algorithm for defending against adversarial attacks, \DEFENSE{}, is detailed in Algorithm~\ref{alg:explainer_defense}.
The approach is a fairly straightforward modification to the neighborhood generation function $\pi_{\mathbf{x}_i}^{\ginline{}}$. For each sample $\mathbf{x}_i$ to be explained by $g$, the perturbed samples $\bm{\mathcal{X}}^{\ginline{}}_i$ generated by $\pi_{\mathbf{x}_i}^{\ginline{}}$ are scored by $h$. The samples with scores below the threshold $h.\tau$ are discarded and \DEFENSE{} recursively builds the remaining samples until $|\bm{\mathcal{X}}^{\ginline{}}_i|=n_p$.
In practice, the recursive depth can be limited with either an explicit limit or by reducing $h.\tau$. However, an auditor will prioritize faithful scrutiny of \ACME{}'s algorithms over the speed of the explainer.

On a final note, neither \DETECT{} nor \DEFENSE{} is tied to \KNNCAD{} --- rather, they are compatible with any $h \in \mathcal{H}$ (any conditional anomaly detector).

\paragraph{Time and Space Complexity}

\begin{table}[t]
    \small
    \centering
    \ra{1.2}
    \addtolength{\tabcolsep}{-2pt}
    \begin{tabular}{@{}ccll@{}}
        \toprule
        Alg. & Type & \makecell{Complexity} & \makecell{Practical\\Complexity}  \\
        \midrule
        \multirow{2}{*}{\ref{alg:knncad_fit}} 
        & Time & $\mathcal{O}(N ((F + k)\log N + T_{f}(F)))$ & $\mathcal{O}(N T_{f}(F))$ \\
        & Space & $\mathcal{O}(N (F + k + S_{f}(F)))$ & $\mathcal{O}(N S_{f}(F))$ \\[1.5ex]
        \multirow{2}{*}{\ref{alg:knncad_score}} 
        & Time & $\mathcal{O}(N (k \log N + T_{f}(F)))$ & $\mathcal{O}(N T_{f}(F))$ \\
        & Space & $\mathcal{O}(N (k + S_{f}(F)))$ & $\mathcal{O}(N S_{f}(F))$ \\[1.5ex]
        \multirow{2}{*}{\ref{alg:adv_detect}} & 
        Time & $\mathcal{O}(N n_p (k \log (N n_p) + T_{f}(F)))$ & $\mathcal{O}(N n_p T_{f}(F))$ \\
        & Space & $\mathcal{O}(N n_p (k + S_{f}(F)))$ & $\mathcal{O}(N n_p S_{f}(F))$ \\[1.5ex]
        \multirow{2}{*}{\ref{alg:explainer_defense}} 
        & Time & $\mathcal{O}(R n_p (k \log n_p + T_{f}(F)))$ & $\mathcal{O}(R n_p T_{f}(F))$ \\
        & Space & $\mathcal{O}(n_p (k + S_{f}(F)))$ & $\mathcal{O}(n_p S_{f}(F))$ \\
        \bottomrule
    \end{tabular}
    \caption{The time and space complexity of each algorithm introduced in this paper: Algorithms~\ref{alg:knncad_fit} (\KNNCAD{}{\small\texttt{.fit}}), \ref{alg:knncad_score} (\KNNCAD{}{\small\texttt{.score\_samples}}), \ref{alg:adv_detect} (\DETECT{}), and \ref{alg:explainer_defense} (\DEFENSE{}). With practical complexity, it is assumed that $T_{f}(F) \gg (k+F) \log N$ and $S_{f}(F) \gg k+F$, \eg{}, as is the case with DNNs and most decision trees.}
    \label{tab:complexity}
    \addtolength{\tabcolsep}{2pt}
\end{table}

The time and space complexity of every introduced algorithm are listed in Table~\ref{tab:complexity} and derived in detail in Appendix~I.
We assume that \KNNCAD{} is used as $h$ in Algorithms~\ref{alg:adv_detect} and~\ref{alg:explainer_defense}. Note that \KNNCAD{} uses the ball tree algorithm in computing nearest neighbors.
We let $T_{f}(\cdot)$ and $S_{f}(\cdot)$ be the functions that give the time and space complexity of $f$, respectively, and $R$ be the number of recursions in Algorithm~\ref{alg:explainer_defense}.
In practice, the time and space complexity of each algorithm are dominated by that of the model to audit, $f$. Hence, reducing the number of queries is desirable. We explore the sample-efficiency of our approach in Section~\ref{sec:experiments}.
If the queries are pre-computed, then our algorithms using \KNNCAD{} take linearithmic time and linear space.

%% file: algs/knncad.tex
\begin{algorithm}[t]%
    \small
    \KwIn{$f$, $\bm{\mathcal{X}}$, $k$, $\phi$, $\epsilon$}
    \KwOut{$h$, the \KNNCAD{} object}
    \tcp{Fit the distribution of normality}
    $h \gets \texttt{KNN}(\bm{\mathcal{X}})$\;
    $\mathcal{S} \gets h.\texttt{score\_samples}(f, \bm{\mathcal{X}}, k, \phi)$\;
    $\mathcal{S}' \gets \texttt{sort}(\mathcal{S})$\tcp*[r]{sort ascending}
    $h.\tau \gets \mathcal{S}'[\text{round}({\epsilon \cdot |\mathcal{S}'|})]$\tcp*[r]{set threshold}
    $h.\bm{\mathcal{X}}_{\text{train}} \gets \bm{\mathcal{X}};~~ h.\bm{\mathcal{Y}}_{\text{train}} \gets f(\bm{\mathcal{X}})$\;
    \KwRet{$h$}
    \caption{\KNNCAD{}{\small\texttt{.fit}}($f, \bm{\mathcal{X}}, k, \phi, \epsilon$)}
    \label{alg:knncad_fit}
\end{algorithm}%
\begin{algorithm}[t]%
    \small
    \KwIn{$h$, $f$, $\bm{\mathcal{X}}$, $k$, $\phi$}
    \KwOut{$\mathcal{S}$, the scored samples}
    $\bm{\mathcal{Y}} \gets f(\bm{\mathcal{X}})$\;
    \tcp{Get the $k$-nearest neighbor distances and indices}
    $\mathcal{D},\, \mathcal{I} \gets h\texttt{.neighbors}(\bm{\mathcal{X}}, k)$\;
    $\mathcal{S} \gets \texttt{new array}$\;
    \For{$(\mathbf{d}, \mathbf{i}, y_j) \in (\mathcal{D}, \mathcal{I}, \bm{\mathcal{Y}})$}{
        $\mathbf{y}_i \gets h.\bm{\mathcal{Y}}_{\text{train}}[\mathbf{i}]$\tcp*[r]{Gather neighbor labels}
        $\mathbf{d}_0 \gets \mathbf{d}[\llbracket \mathbf{y}_i = 0 \rrbracket];~~ \mathbf{d}_1 \gets \mathbf{d}[\llbracket \mathbf{y}_i = 1 \rrbracket]$\;
        $d_0 \gets \phi(\mathbf{d}_0);~~ d_1 \gets \phi(\mathbf{d}_1)$\;
        \eIf{$y_j = 1$}{
            $d_{y_j} \gets d_1;~~ d_{\neg{}y_j} \gets d_0$\;
        }{
            $d_{y_j} \gets d_0;~~ d_{\neg{}y_j} \gets d_1$\;
        }
        $s \gets \zeta(d_{\neg{}y_j}, d_{y_j})$\;
        $\mathcal{S}\texttt{.append}(s)$\;
    }
    \KwRet{$\mathcal{S}$}
    \caption{\KNNCAD{}{\small\texttt{.score\_samples}}($h, f, \bm{\mathcal{X}}, k, \phi$)}
    \label{alg:knncad_score}
\end{algorithm}%
\begin{algorithm}[t]%
    \small
    \KwIn{$f$, $g$, $h$, $\Theta_h$, $\bm{\mathcal{X}}$, $n_{\text{train}}$, $\tau_{\text{global}}$}
    \KwOut{Whether $f$ is behaving adversarially toward $g$}
    $\bm{\mathcal{X}}_{\text{train}}, \bm{\mathcal{X}}_{\text{test}} \gets \bm{\mathcal{X}}[{:}n_{\text{train}}], \bm{\mathcal{X}}[n_{\text{train}}{:}]$\;
    $h\texttt{.fit}(f, \bm{\mathcal{X}}_{\text{train}}, \Theta_h)$\;
    $\mathcal{S}_{\text{test}} \gets h\texttt{.score\_samples}(f, \bm{\mathcal{X}}_{\text{test}}, \Theta_h)$\;
    $\bm{\mathcal{X}}^{(g)}_{\text{test}} \gets \{\}$\;
    \For{$\mathbf{x}_i \in \bm{\mathcal{X}}_{\textup{test}}$}{
        $\bm{\mathcal{X}}^{(g)}_{\text{test}} \gets \bm{\mathcal{X}}^{(g)}_{\text{test}} \cup \pi^{(g)}_{\mathbf{x}_i}(\mathbf{x}_i)$\;
    }
    $\mathcal{S}_{\text{test}}^{(g)} \gets h\texttt{.score\_samples}(f, \bm{\mathcal{X}}_{\text{test}}^{(g)}, \Theta_h)$\;
    $a_{\text{test}} \gets \texttt{AUC}(\texttt{ECDF}(\mathcal{S}_{\text{test}}))$\;
    $a_{\text{test}}^{(g)} \gets \texttt{AUC}(\texttt{ECDF}(\mathcal{S}_{\text{test}}^{(g)}))$\;
    \KwRet{$\llbracket (a_{\textup{test}}^{(g)} - a_{\textup{test}}) \ge \tau_{\textup{global}} \rrbracket$}
    \caption{\DETECT{}($f, g, h, \Theta_h, \bm{\mathcal{X}},n_{\textup{train}},\tau_{\textup{global}}$)}
    \label{alg:adv_detect}
\end{algorithm}%
\begin{algorithm}[t]%
    \small
    \KwIn{$f$, $g$, $h$, $\mathbf{x}_i$, $\Theta_h$, $n_{p}$}
    \KwOut{$\bm{\mathcal{X}}^{(g)}_i$, (more) in-distribution perturbations}
    $\bm{\mathcal{X}}^{(g)}_i \gets \pi^{(g)}_{\mathbf{x}_i}(\mathbf{x}_i)$\;
    $\mathcal{S}^{(g)}_i \gets h\texttt{.score\_samples}(f, \bm{\mathcal{X}}^{(g)}_i, \Theta_h)$\;
    \tcp{Remove abnormal samples}
    $\bm{\mathcal{X}}^{(g)}_i \gets \bm{\mathcal{X}}^{(g)}_i[\llbracket \mathcal{S}^{(g)}_i > h.\tau \rrbracket]$\;
    $n'_{p} \gets n_{p}-|\bm{\mathcal{X}}^{(g)}_i|$\;
    \If{$n'_{p} \ne 0$}{
        $\bm{\mathcal{X}}'^{(g)}_i \gets \DEFENSE{}(f, g, h, \mathbf{x}_i, \Theta_h, n'_{p})$\;
        $\bm{\mathcal{X}}^{(g)}_i \gets \bm{\mathcal{X}}^{(g)}_i \cup \bm{\mathcal{X}}'^{(g)}_i$\;
    }
    \KwRet{$\bm{\mathcal{X}}^{(g)}_i$}
    \caption{\DEFENSE{}($f, g, h, \mathbf{x}_i, \Theta_h,n_{p}$)}
    \label{alg:explainer_defense}
\end{algorithm}%

%% file: src/05_Experiments.tex
\section{Experiments}\label{sec:experiments}

We consider three real-world high-stakes data sets to evaluate our approach:
\begin{itemize}[noitemsep,nolistsep,leftmargin=1em]
    \item The Correctional Offender Management Profiling for Alternative Sanctions~(\textbf{COMPAS}) dataset was collected by \textit{ProPublica} in 2016 for defendants from Broward County, Florida~\cite{angwin2016machine}. The attributes of individuals are used by the COMPAS algorithm to assign recidivism risk scores provided to relevant decision-makers. %
    \item The \textbf{German Credit} data set, donated to the University of California Irvine (UCI) machine learning repository in 1994, comprises a set of attributes for German individuals and the corresponding lender risk~\cite{Dua:2019}.
    \item The \textbf{Communities and Crime} data set combines socio-economic US census data (1990), US Law Enforcement Management and Administrative Statistics (LEMAS) survey data (1990), and US FBI Uniform Crime Reporting (UCR) data (1995)~\cite{REDMOND2002660}. Covariates describing individual communities are posited to be predictive of the crime rate.
\end{itemize}

Each data set contains at least one protected attribute that should not be used to make a decision in order to meet regulatory
compliance.
We follow the attacks as proposed in~\cite{fooling} and recall them here.
See Table~\ref{tab:datasets} for the sensitive features that each $f_{\text{biased}}$ uses and the harmless features that each $f_{\text{unbiased}}$ uses to make decisions for each experiment.
As in~\cite{fooling}, the OOD detector $\dood{}$ is selected to be a random forest.
Uncorrelated features are generated from the discrete uniform distribution $\mathcal{U}\{0, 1\}$, and two different $f_{\text{unbiased}}$ are considered for each experiment when uncorrelated features are considered: one with one uncorrelated feature and another with two uncorrelated features.
The two classifiers, $f_{\text{biased}}$ and $f_{\text{unbiased}}$, are rule-based and return the target listed in Table~\ref{tab:datasets} as a function of the selected features(s). See Appendix~A for further details.

\begin{table}[t]
    \small
    \centering
    \ra{1.5}
    \begin{tabular}{@{}lccc@{}}
        \toprule
        {Data Set} & \makecell{Sensitive\\Feature} & \makecell{Harmless\\Feature(s)} & {Target} \\
        \midrule
        {COMPAS} & \makecell{African-\\American} & Uncorrelated & \makecell{High Risk of\\Recidivism} \\
        \makecell[l]{German\\Credit} & Gender & \makecell{Income \% \\Toward Loan} & Good Customer \\
        {CC} & \makecell{Count of White\\Population} & Uncorrelated & \makecell{Violent\\Crime Rate} \\
        \bottomrule
    \end{tabular}
    \caption{Data sets on which the attack and defense approaches are evaluated. The Communities \& Crime data set is abbreviated as CC. In each experiment, $f_{\text{biased}}$ makes decisions based on sensitive features and $f_{\text{unbiased}}$ makes decisions based on harmless features.}
    \label{tab:datasets}
\end{table}

For each task, we consider both cases in which \ACME{} elects to and not to employ the attack.
Each data set contains a train and test partition, $\bm{\mathcal{D}}_{\text{train}}$ and $\bm{\mathcal{D}}_{\text{test}}$, respectively --- both are considered to be representative of normality (as defined in Section~\ref{sec:methods}).
$f$ is trained on $\bm{\mathcal{D}}_{\text{train}}$ and scrutinized by the auditor on $\bm{\mathcal{D}}_{\text{test}}$ using $g$.
Recall that $f$ is a function of both $f_{\text{biased}}$ and $f_{\text{unbiased}}$ as defined in \eqref{eq:adv_scaffold}.
Our defense approach aided by $h$ employs the algorithms \DETECT{} and \DEFENSE{}. The defense is evaluated using $\bm{\mathcal{D}}_{\text{test}}$ as the auditor does not have access to $\bm{\mathcal{D}}_{\text{train}}$.
The results across all tasks are shown in Table~\ref{tab:results}.
Appendix~B specifies all reproducibility details. In the proceeding sections, we propose and employ several fidelity metrics to evaluate the quality of both the attack and defense.

\begin{figure*}
    \centering
    \begin{subfigure}[b]{0.48\textwidth}
        \centering
        \includegraphics[width=\linewidth]{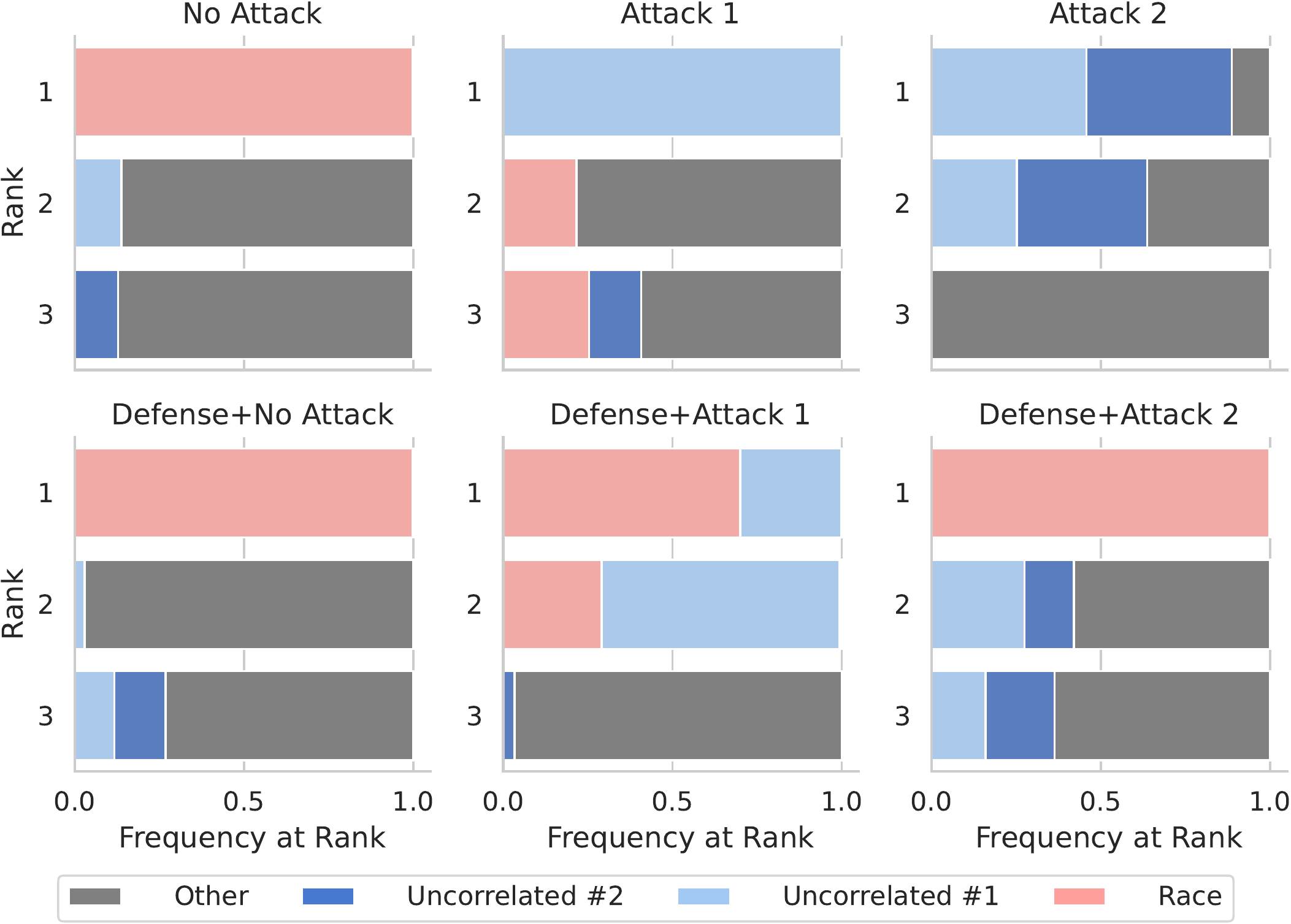}
    \end{subfigure}\hfill
    \begin{subfigure}[b]{0.48\textwidth}
        \centering
        \includegraphics[width=\linewidth]{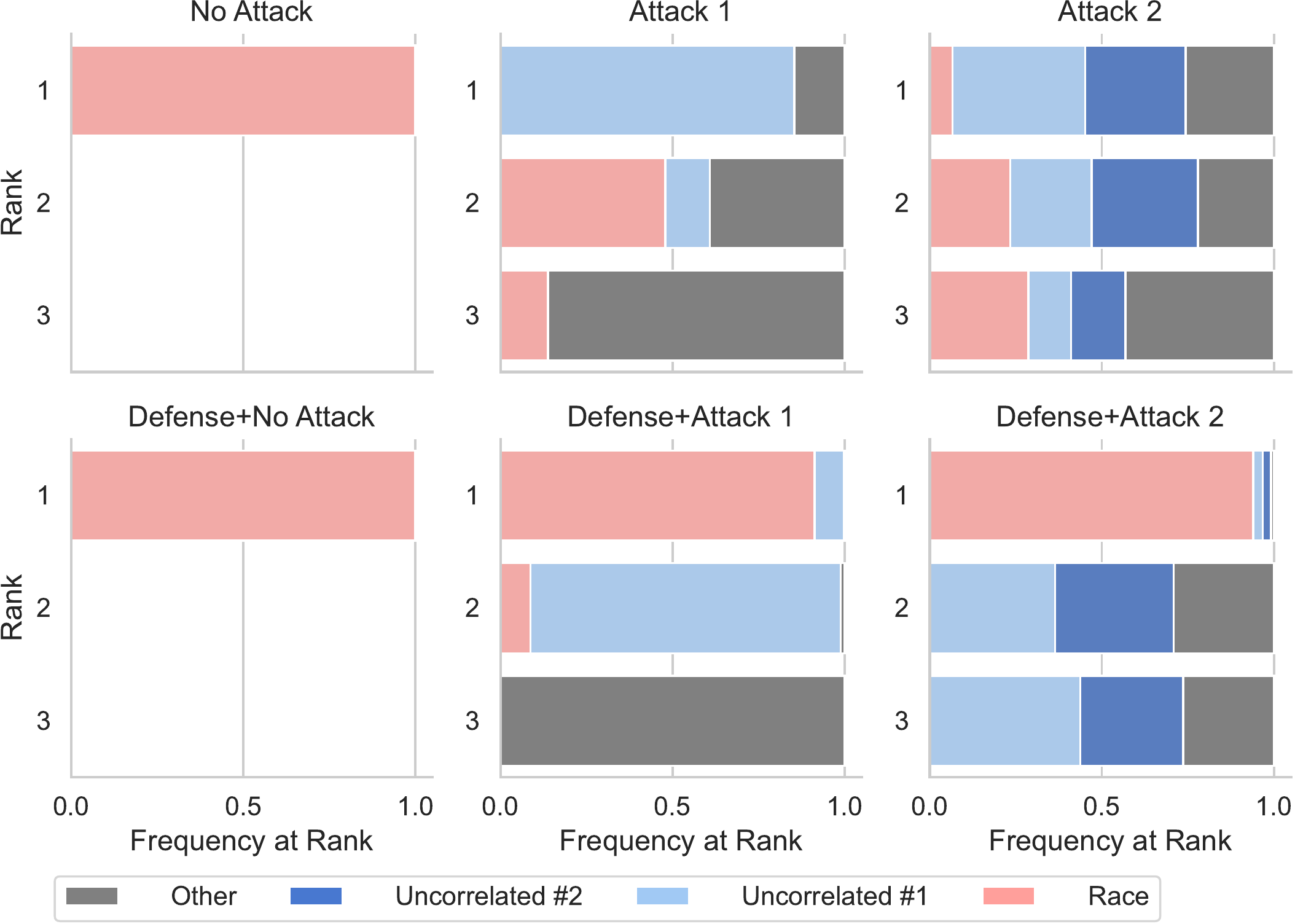}
    \end{subfigure}
    \caption{The top-3 most important features identified by \LIME{} (left) and \SHAP{} (right) with and without the defense on the COMPAS task. Our defense mitigates the attack and demonstrates that \ACME{} attempted to conceal that race is an important factor in deciding the risk of recidivism. In the legend, ``Other'' represents any other COMPAS feature that is not ``Race'' or an ``Uncorrelated'' feature.}
    \label{fig:topk}
\end{figure*}

\paragraph{How effective is the attack?}
We evaluate the efficacy of the attack with several metrics, expanding on its original evaluation~\cite{fooling}.
The classifier $f$ should be accurate on in-distribution data ($\text{fidelity}_{f}$) as given by \eqref{eq:f_fidelity}
\begin{equation}\label{eq:f_fidelity}
    \text{fidelity}_{f}(\bm{\mathcal{X}}) = \frac{1}{|\bm{\mathcal{X}}|} \sum_{\mathbf{x}_i \in \bm{\mathcal{X}}} \llbracket f(\mathbf{x}_i) = f_{\text{actual}}(\mathbf{x}_i) \rrbracket 
\end{equation}
where
\begin{equation}\label{eq:factual}
    f_{\text{actual}}(\mathbf{x}_i) =
    \begin{cases}
        f_{\text{biased}}(\mathbf{x}_i)    & \text{if attack is deployed} \\
        f(\mathbf{x}_i) & \text{otherwise}
    \end{cases}
\end{equation}
When the attack is deployed, we can evaluate the quality of $\dood{}$ ($\text{fidelity}_{\dood{}}$) as the balanced accuracy between samples from $\bm{\mathcal{X}}$ and $\bm{\mathcal{X}}^{\ginline{}}$ as in \eqref{eq:dood_fidelity}
\begin{align}\label{eq:dood_fidelity}
\begin{split}
&    \text{fidelity}_{\dood{}}(\bm{\mathcal{X}} \cup \bm{\mathcal{X}}^{(g)}) =\\&
    \frac{1}{2}
    \hspace{-1pt}\bigg(
    \frac{1}{|\bm{\mathcal{X}}|}
    \hspace{-1pt}\sum_{\mathbf{x}_i \in \bm{\mathcal{X}}}
    \hspace{-1pt}\llbracket \dood{}(\mathbf{x}_i) {=} 1 \rrbracket {+}
    \frac{1}{|\bm{\mathcal{X}}^{(g)}|}
    \hspace{-1pt}\sum_{\mathbf{x}_i \in \bm{\mathcal{X}}^{(g)}}
    \hspace{-3pt}\llbracket \dood{}(\mathbf{x}_i) {=} 0 \rrbracket
    \hspace{-1pt}\bigg).
\end{split}
\end{align}
To quantify the degree that $g$ is ``fooled,'' we compare the feature contributions to the features used by $f_{\text{actual}}$ from a feature-selectivity perspective.
The features used by $f_{\text{actual}}$ can be split into selected ($\mathcal{F}_{\text{selected}}$) and unselected ($\mathcal{F}_{\neg{}\text{selected}}$) features.
In turn, we can estimate the fidelity of $g$ as~\eqref{eq:g_fidelity}
\begin{align}\label{eq:g_fidelity}
\begin{split}
    \text{fidelity}_{g}(\bm{\mathcal{X}}) {=}
    \frac{1}{F {\cdot} |\bm{\mathcal{X}}|}
    \sum_{\mathbf{x}_i {\in} \bm{\mathcal{X}}}
        \rho\big(&
        \{
            F{-}1 {\mid} j {\in} \mathcal{F}_{\text{selected}}
        \} {\cup}\\&
        \{
            0 {\mid} j {\in} \mathcal{F}_{\neg{}\text{selected}}
        \},\\&
        \{
            r(\mathcal{E}_i^{-}, a_{ij}) {\mid} j {\in} \mathcal{F}
        \}
        \big)
\end{split}
\end{align}
where
$\mathcal{F} = \mathcal{F}_{\text{selected}} \cup \mathcal{F}_{\neg{}\text{selected}}$, $|\mathcal{F}|{=}F$,
and $\rho$ gives the Spearman rank correlation coefficient between its two ordered arguments. Put simply, the correlation increases when features in $\mathcal{F}_{\text{selected}}$ are ranked higher than those in $\mathcal{F}_{\neg{}\text{selected}}$ by an explainer, \ie{}, explanations are more faithful.

As shown in Table~\ref{tab:results}, the attack is quite effective --- fidelity$_{f}$ remains high when the attack is active while fidelity$_{g}$ decreases. This indicates that $\dood{}$ successfully toggles between $f_{\text{biased}}$ and $f_{\text{unbiased}}$, which is supported by the fidelity$_{\dood{}}$ scores. Figure~\ref{fig:topk} qualitatively shows the efficacy of the attack on the COMPAS data set by visualizing the top-3 most frequent features in explanations.

\input{tables/results}

\paragraph{Can we detect the attack?}
We introduce metrics to evaluate $h$ and \DETECT{} in their ability to detect attacks.
Here, $d_{\text{actual}}$ is $\dood{}$ when a scaffolding attack is deployed and a dummy function that returns `1' otherwise.
The ability of $h$ to model $\mathbb{P}(f(\mathbf{x}_i) {\mid} \mathbf{x}_i)$ is given by the weighted mean-square-error of $h$ and $d_{\text{actual}}$ (fidelity$_{h}$) as in~\eqref{eq:h_fidelity}
\begin{align}\label{eq:h_fidelity}
\begin{split}
    \text{fidelity}_{h}&(\bm{\mathcal{X}} {\cup} \bm{\mathcal{X}}^{(g)}) =\\
    1 -
    \frac{1}{2}\bigg(&
    \frac{1}{|\bm{\mathcal{X}}|}
    \sum_{\mathbf{x}_i \in \bm{\mathcal{X}}}
    \left(d_{\text{actual}}(\mathbf{x}_i) {-} h(\mathbf{x}_i)\right)^2 +\\&
    \frac{1}{|\bm{\mathcal{X}}^{(g)}|}
    \sum_{\mathbf{x}_i \in \bm{\mathcal{X}}^{(g)}}
    \left(d_{\text{actual}}(\mathbf{x}^{(g)}_i) {-} h(\mathbf{x}^{(g)}_i)\right)^2\bigg)
\end{split}
\end{align}
\noindent
where $h(\cdot)$ is shorthand for $h${\small\texttt{.score\_samples}}$(\{\cdot\})$.
Since $h$ gives probabilities, the error term is bounded to the range $[0,1]$ which we subtract from `1' in order for the score to be interpreted as fidelity.
To evaluate the quality of \DETECT{}, we compare the areas between the cumulative distributions of scores derived from $\bm{\mathcal{X}}^{\ginline{}}$ and $\bm{\mathcal{X}}$ as described in Algorithm~\ref{alg:adv_detect}: $\Delta_{\text{cdf}} := a_{\text{test}}^{\ginline{}} - a_{\text{test}}$. This score should be higher if there is an attack deployed and lower otherwise.

Across all experiments, we use \KNNCAD{} as $h$.
Our approach is able to detect every attack without false positives when $\tau_{\text{global}} \in [0.00,0.12]$ for \SHAP{} and $\tau_{\text{global}} \in [0.11,0.12]$ for \LIME{}.
There is a trade-off in fidelity$_h$ and fidelity$_{\dood{}}$ --- the more off-manifold the perturbations are, the harder it is to fit
$\mathbb{P}(f(\mathbf{x}_i) \mid \mathbf{x}_i)$
(without an attack), whereas it is harder to fit $\dood{}$ when the perturbations are more on-manifold.
The more principled game-theoretic perturbation scheme of \SHAP{} demonstrably improves on that of \LIME{}, yet it still can be gamed adversarially.
Furthermore, a higher $\Delta_{\text{cdf}}$ without an attack deployed is indicative of more perturbed samples being off-manifold --- discriminating attacks using \SHAP{} has a far greater margin between $\Delta_{\text{cdf}}$ scores than \LIME{}.

\paragraph{Can we ``unfool'' explainers?}
In order for an explainer to be ``unfooled,'' it should produce explanations that 1) agree with its own explanations without an attack present and 2) agree with the feature selection by the black box algorithm.
We propose $\text{infidelity}_{\DEFENSE{},g}$ to quantify how well \DEFENSE{} accomplishes the former goal. This metric measures the disagreement of the explanations when using \DEFENSE{} to those only using $g$ as in \eqref{eq:defense_original_explanation_fidelity}
\begin{align}\label{eq:defense_original_explanation_fidelity}
\begin{split}
    \text{infi}&\text{delity}_{\DEFENSE{},g}(\bm{\mathcal{X}}) =\\
    &\frac{1}{F\cdot |\bm{\mathcal{X}}|} \sum_{\mathbf{x}_i \in \bm{\mathcal{X}}}
    \sum_{(a^{(g)}_{ij}, a^{(h)}_{ij})\in (\mathcal{E}^{(g)}_i, \mathcal{E}^{(h)}_i)}
    (a^{(g)}_{ij} - a^{(h)}_{ij}) ^ 2
\end{split}
\end{align}
where $\mathcal{E}^{\ginline{}}_i$ is the explanation from $g$ without an attack, $\mathcal{E}^{\hinline{}}_i$ is the explanation when $h$ is used for defense during an attack, and $a_{ij}^{(*)}$ is as defined in Section~\ref{sec:background}.
To evaluate against the true feature selection of $f_{\text{actual}}$, we use the same metric definition as $\text{fidelity}_{g}$ and refer to it as $\text{fidelity}_{\DEFENSE{},f}$.

The
infidelity
scores in Table~\ref{tab:results} are near-zero across all tasks, demonstrating that our defense has very little disagreement with the explainer $g$ when $g$ is not under attack.
In addition, the $\text{fidelity}_{\DEFENSE{},f}$ scores exceed those of $\text{fidelity}_{g}$ for all experiments when an attack is deployed and are close to $\text{fidelity}_{g}$ without an attack.
A final piece of evidence that our defense mitigates the attack is shown in Figure~\ref{fig:topk} --- comparing the top-3 most important features with and without the defense confirms that the defense is highly successful. See Appendix~E for the same figure with the remaining explainers and data sets.

\begin{figure}[]
    \centering
    \includegraphics[width=\linewidth]{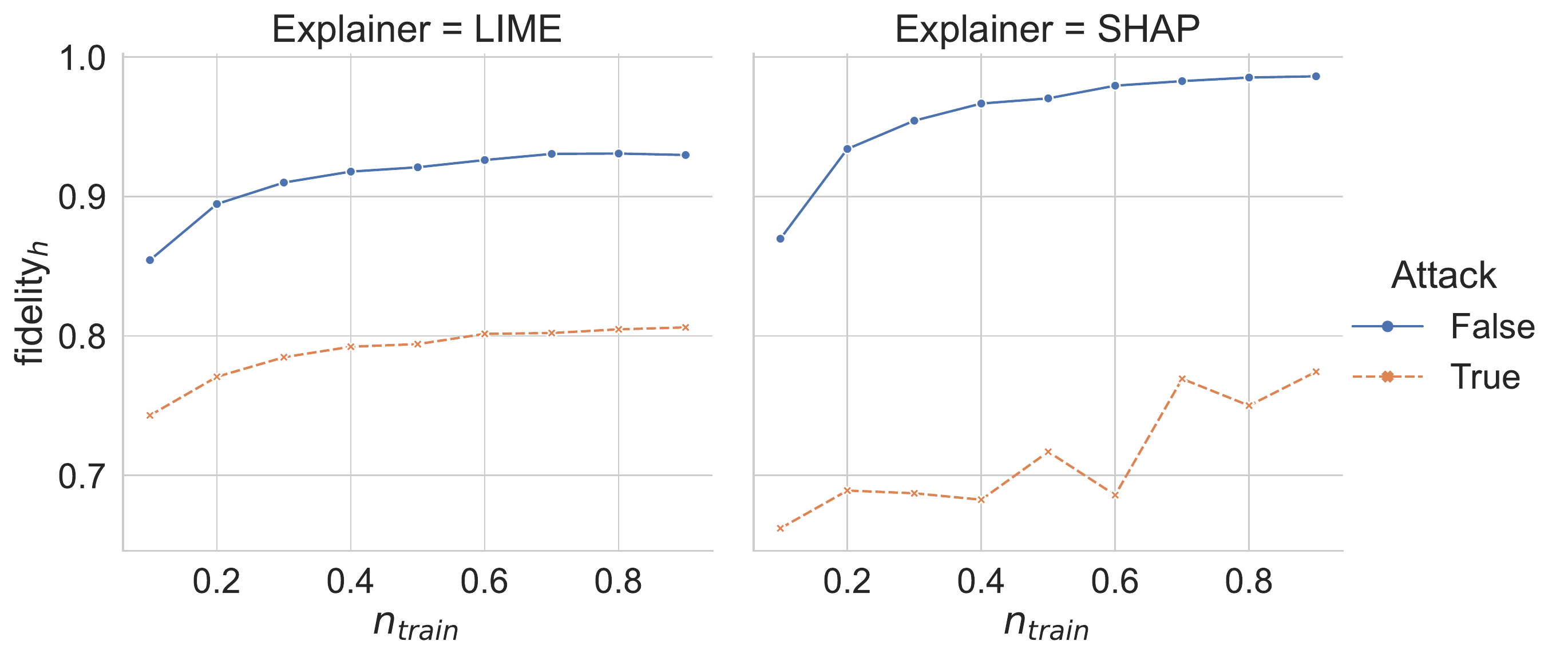}
    \caption{Sample-efficiency of \KNNCAD{} on the COMPAS data set when using \LIME{} and \SHAP{} as $g$ --- fidelity$_h$ is plotted as a function of the training proportion of the COMPAS data ($n_{\text{train}}$).}
    \label{fig:sample_eff_h}
\end{figure}

\begin{figure}[]
    \centering
    \includegraphics[width=\linewidth]{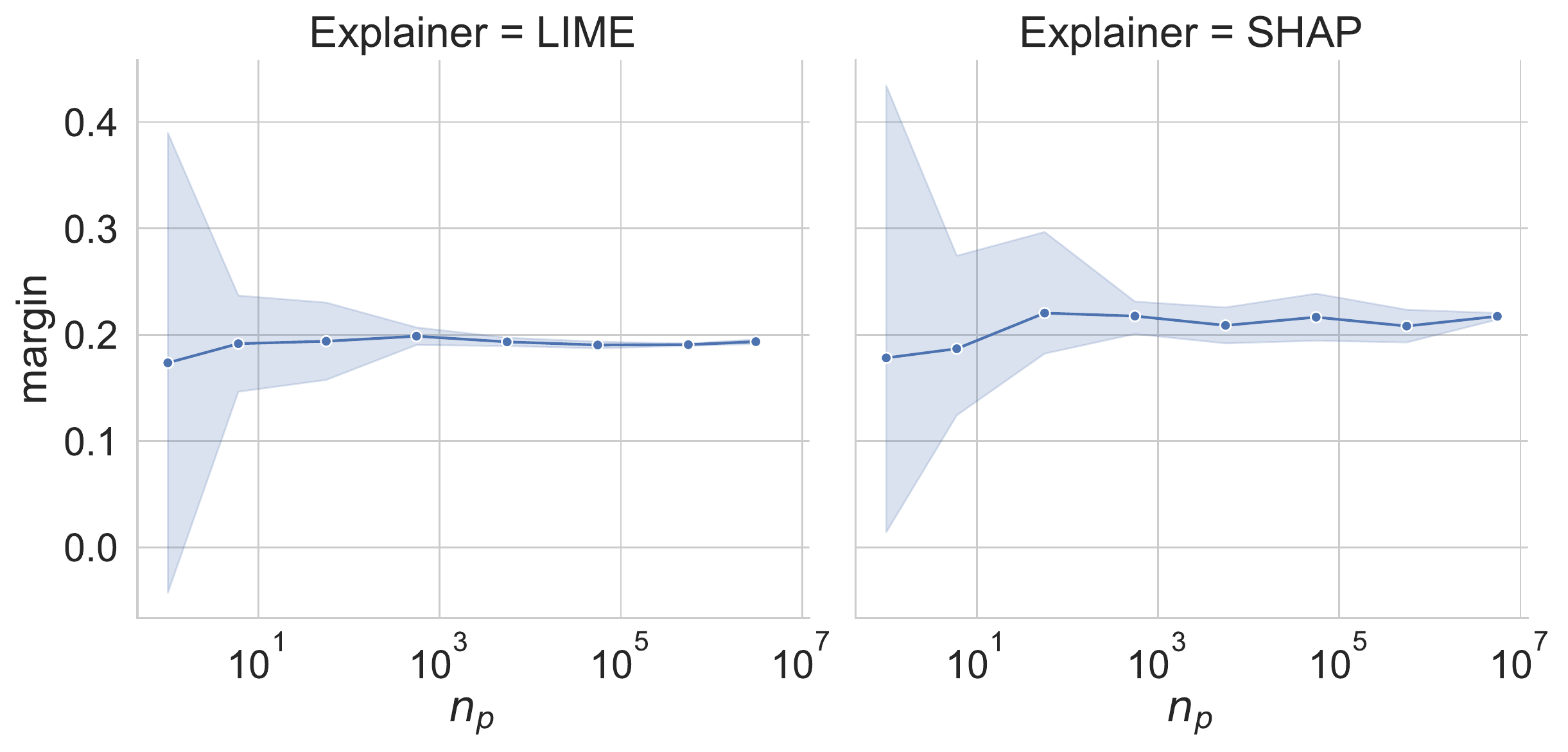}
    \caption{Sample-efficiency of \DETECT{} on the COMPAS data set when using \LIME{} and \SHAP{} as $g$ --- the margin between the $\Delta_{cdf}$ scores with and without an attack is plotted as a function of the number of explainer perturbations ($n_{p}$).}
    \label{fig:sample_eff_detect}
\end{figure}

\paragraph{Sample Efficiency}

We evaluate the training sample-efficiency of our approach. Figure~\ref{fig:sample_eff_h} plots fidelity$_h$ as a function of the training proportion of the COMPAS data when using \LIME{} and \SHAP{} as $g$. Because of the strong inductive bias and nonparametric nature of \KNNCAD{}, the algorithm hardly degrades in performance even when 10\% of the data is in use.
This in part indicates the sample-efficiency of the \DETECT{} and \DEFENSE{} algorithms when \KNNCAD{} is used.
We also evaluate the sample-efficiency of \DETECT{} on the COMPAS data set when using \LIME{} and \SHAP{} as $g$. The margin between the $\Delta_{\text{cdf}}$ scores with and without an attack is plotted as a function of the number of explainer perturbations, $n_{p}$. The main benefit of increasing $n_p$ is to increase the consistency of the detection score. When $n_p > 1,000$ the variance tapers off for both explainers, which is just a small percentage of the millions of explainer-generated perturbations when all test set samples are explained.
These findings are quite notable as the complexity bottleneck is due to querying $f$ as discussed in Section~\ref{sec:methods}.

\paragraph{Additional Analyses}
We include analyses of the hyperparameters of the three core algorithms, \KNNCAD{}, \DETECT{} and \DEFENSE{}, in Appendix~F.

%% file: tables/results.tex
\begin{table*}%
    \small
    \centering
    \ra{1.1025}
    \begin{tabular}{@{}lcccccccccccccccc@{}}
        \toprule
        \multirow{4}{*}{\rotatebox{0}{Data Set}} & \multirow{4}{*}{Attack} & \multirow{4}{*}{${N_{hl}}$} &
        \multicolumn{7}{c}{$g=$ \LIME{}} & \multicolumn{7}{c}{$g=$ \SHAP{}} \\
        \cmidrule(lr){4-10}\cmidrule(l){11-17}
        &&&\multicolumn{4}{c}{Fidelity}
        & {\DETECTtiny{}} & \multicolumn{2}{c}{\DEFENSEtiny{}}
        &\multicolumn{4}{c}{Fidelity}
        & {\DETECTtiny{}} & \multicolumn{2}{c}{\hspace{1.75ex}\DEFENSEtiny{}} \\
        \cmidrule(lr){4-7}\cmidrule(lr){8-8}\cmidrule(lr){9-10}\cmidrule(lr){11-14}\cmidrule(lr){15-15}\cmidrule(l){16-17}
        &&& ${f}$ & ${\dood{}}$ & ${g}$ & ${h}$ & $\Delta_{\text{cdf}}$ & $\text{inf}_{g}$ & $\text{fid}_{f}$
        & ${f}$ & ${\dood{}}$ & ${g}$ & ${h}$ & $\Delta_{\text{cdf}}$ & $\text{inf}_{g}$ & $\text{fid}_{f}$ \\
        \midrule
        \multirow{3}{*}{COMPAS}%
        &\xmark& --
        & 1.00 & --   & 0.31 & 0.93 & 0.11 & 0.00 & 0.31
        & 1.00 & --   & 0.29 & 0.99 & -0.02 & 0.00 & 0.29 \\
        &\cmark& 1
        & 0.99 & 0.99 & 0.23 & 0.81 & 0.30 & 0.05 & 0.30
        & 0.93 & 0.82 & 0.24 & 0.77 & 0.20 & 0.02 & 0.28 \\
        &\cmark& 2
        & 0.99 & 0.99 & 0.10 & 0.80 & 0.30 & 0.00 & 0.31
        & 0.91 & 0.82 & 0.24 & 0.77 & 0.25 & 0.01 & 0.26 \\[1.25ex]
        \multirow{2}{*}{\makecell[l]{German \\ Credit}}%
        &\xmark& --
        & 1.00 & --   & 0.27 & 0.78 & 0.07 & 0.00 & 0.27
        & 1.00 & --   & 0.24 & 0.86 & -0.05 & 0.00 & 0.27 \\
        &\cmark& 1
        & 1.00 & 0.99 & 0.18 & 0.70 & 0.18 & 0.00 & 0.27
        & 0.83 & 0.71 & 0.12 & 0.61 & 0.12 & 0.00 & 0.27 \\[1.25ex]
        \multirow{3}{*}{CC}%
        &\xmark& --
        & 1.00 & --   & 0.21 & 0.80 & 0.09 & 0.00 & 0.16
        & 1.00 & --   & 0.08 & 0.85 & 0.00 & 0.00 & 0.20 \\
        &\cmark& 1
        & 1.00 & 0.99 & 0.16 & 0.79 & 0.12 & 0.01 & 0.20
        & 0.97 & 0.85 & 0.07 & 0.79 & 0.13 & 0.01 & 0.19 \\
        &\cmark& 2
        & 1.00 & 0.99 & 0.17 & 0.79 & 0.12 & 0.00 & 0.31
        & 0.99 & 0.85 & 0.07 & 0.77 & 0.12 & 0.00 & 0.30 \\
        \bottomrule
    \end{tabular}
    \caption{The (in)fidelity and $\Delta_{\text{cdf}}$ scores of the attack and defense across the real-world data sets when \KNNCAD{} is used as $h$.
    $N_{hl}$ is the number of harmless features used by $f_{\text{unbiased}}$ and is unspecified when no attack is deployed. See the text for the differing definitions of fidelity for each algorithm. We abbreviate $\text{infidelity}_{\DEFENSE{},g}$ as $\text{inf}_{g}$ and $\text{fidelity}_{\DEFENSE{},f}$ as $\text{fid}_{f}$. Our approach successfully detects the scaffolding attack while maintaining explanation fidelity for both explainers.}
    \label{tab:results}
\end{table*}

%% file: src/06_Discussion.tex
\section{Discussion}\label{sec:discussion}

In this work, we introduced several novel algorithms to defend against adversarial attacks on perturbation-based \posthoc{} explainers: \KNNCAD{} for conditional anomaly detection, \DETECT{} for attack detection, and \DEFENSE{} to improve the fidelity of explanations when under attack. We rigorously formalized the attack and defense models, as well as introduced new quantitative metrics to evaluate the quality of the attack and defense. Our approach demonstrably mitigated the scaffolding attack across several real-world high-stakes data sets. The results indicate that the defense of \SHAP{} is easier than \LIME{} due to
its more realistic data perturbations.

A limitation to consider is that the realistic samples used in the training set for the defense algorithm can be expensive or difficult to collect.
Moreover, in realistic scenarios, an API to a black box may be costly to query with the perturbed samples generated by explainers. In practice, explainer queries should be rate-limited so as to not arise suspicion from the auditee.
On this note, we do not consider the case when the adversary irregularly deploys the attack. We point to \cite{deceptiveAIExpl2022} which characterizes this attack and demonstrates that explanations that are infrequently manipulated can be difficult to detect.
In addition, we considered the case of an adversary masking malicious behavior. However, the motivation for such behavior could arise for privacy reasons or to protect intellectual property from model extraction attacks~\cite{modelExtractionAttack}.

We analyze and alleviate a single shortcoming of \posthoc{} explainers. However, the explainers we consider have also been shown to be inconsistent, unfaithful, and intractable~\cite{krishna2022disagreement,bordt2022posthoc,DBLP:conf/aaai/BroeckLSS21,pmlr-v108-garreau20a,carmichaelposthoceval2021,carmichaelposthoceval2021_2}.
Consequently,
a potential source of negative societal impact in this work arises from practitioners overtrusting \posthoc{} explainers~\cite{kaurInterpretingInterpretabilityUnderstanding2020}. %
Nevertheless, our study demonstrates that the explainers backed with our proposed defense not only detect adversarial behavior but also faithfully identify the most important features in decisions. Moreover, if an explainer is not to be trusted, our approach can at least exploit it to identify misbehaving algorithms.
In future work, the ramifications of an adversarial organization caught red-handed should be explored in the context of existing regulatory guidelines.

%% file: src/07_Acknowledgements.tex
We thank Derek Prijatelj\footnote{\url{https://prijatelj.github.io}} for helpful discussions in the early stages of the conditional anomaly detection formalism. Funding for this work comes from the University of Notre Dame.

%% file: src/99_Supplemental.tex
\renewcommand\thefigure{\thesection\arabic{figure}}
\renewcommand\thetable{\thesection\arabic{table}}
\setcounter{figure}{0}\setcounter{table}{0}
\doparttoc %
\faketableofcontents %
\part{} %
\parttoc %

We highly recommend that this supplemental material be viewed digitally.

\section{Selection of \texorpdfstring{$f_{\text{biased}}$ and $f_{\text{unbiased}}$}{f biased and f unbiased}}

The following details the specifics of the rule-based $f_{\text{biased}}$ and $f_{\text{unbiased}}$ classifiers across the real-world data sets.

\begin{itemize}
    \item COMPAS
    \begin{enumerate}
        \item $f_{\text{biased}}$: high-risk if race is African American, otherwise low-risk. $f_{\text{unbiased}}$: high-risk if uncorrelated\_feature\_1 is 1, otherwise low-risk.
        \item $f_{\text{biased}}$: high-risk if race is African American. $f_{\text{unbiased}}$: high-risk if uncorrelated\_feature\_1 \texttt{XOR} uncorrelated\_feature\_2 is 1, otherwise low-risk.
    \end{enumerate}
    \item German Credit
    \begin{enumerate}
        \item $f_{\text{biased}}$: good customer if gender is male, otherwise bad customer. $f_{\text{unbiased}}$: good customer if LoanRateAsPercentOfIncome is greater than the mean of LoanRateAsPercentOfIncome in the training set, otherwise bad customer.
    \end{enumerate}
    \item Communities \& Crime
    \begin{enumerate}
        \item $f_{\text{biased}}$: low violent crime rate if racePctWhite is above average, otherwise high. $f_{\text{unbiased}}$: low violent crime rate if uncorrelated\_feature\_1, otherwise high.
        \item $f_{\text{biased}}$: low violent crime rate if racePctWhite is above average, otherwise high. $f_{\text{unbiased}}$: low violent crime rate if (uncorrelated\_feature\_1 is greater than 0.5) \texttt{XOR} (uncorrelated\_feature\_2 is greater than -0.5), otherwise high.
    \end{enumerate}
\end{itemize}

\section{Reproducibility}

The following details the hyperparameters, hardware and software used across all experiments.
See Appendix~\ref{sec:error_bars} for a discussion of nondeterminism in experiments and results with error bars over multiple trials. The code is available at the public repository listed in the manuscript.

\paragraph{Hyperparameters}

The hyperparameters used for the $\dood{}$, explainers, and defense algorithms are shown in Table~\ref{tab:dood_hparams}, Table~\ref{tab:explainer_hparams}, and Table~\ref{tab:defense_hparams}, respectively.
All data is normalized using standard z-score normalization based on the mean and standard deviation of the training split of data.

\begin{table}[H]
    \centering
    \begin{tabular}{@{}lr@{}}
        \toprule
        Hyperparameter & Value \\
        \midrule
        \texttt{n\_estimators} & 100 \\
        \texttt{criterion} & \texttt{gini} \\
        \texttt{min\_samples\_split} & \texttt{None} \\
        \texttt{min\_samples\_leaf} & 2 \\
        \texttt{min\_weight\_fraction\_leaf} & 1 \\
        \texttt{max\_features} & \texttt{auto} \\
        \texttt{max\_leaf\_nodes} & \texttt{None} \\
        \texttt{min\_impurity\_decrease} & 0.0 \\
        \texttt{bootstrap} & \texttt{True} \\
        \texttt{oob\_score} & \texttt{False} \\
        \texttt{max\_samples} & \texttt{None} \\
        \bottomrule
    \end{tabular}
    \caption{Hyperparameters of $\dood{}$ used across all experiments. In general, the default \textit{scikit-learn} hyperparameter values are selected. See the \textit{scikit-learn} documentation of \texttt{RandomForestClassifier} for details of each hyperparameter\protect\footnotemark{}.
    }
    \label{tab:dood_hparams}
\end{table}
\footnotetext{\href{https://scikit-learn.org/stable/modules/generated/sklearn.ensemble.RandomForestClassifier.html}{\url{https://scikit-learn.org/stable/modules/generated/sklearn.ensemble.RandomForestClassifier.html}}}

\begin{table}[H]
    \small
    \centering
    \begin{tabular}{@{}lcr@{}}
        \toprule
        Explainer & Hyperparameter & Value \\
        \midrule
        \multirow{6}{*}{\LIME{}}
        & \texttt{kernel\_width} & $F \times 0.75$ \\
        & \texttt{kernel} & \texttt{exponential} \\
        & \texttt{feature\_selection} & \texttt{auto} \\
        & \texttt{num\_features} & \texttt{10} \\
        & \texttt{num\_samples} & \texttt{5000} \\
        & \texttt{discretize\_continuous} & \texttt{False} \\
        \midrule
        \multirow{3}{*}{\SHAP{}}
        & \texttt{background\_distribution} & \texttt{k-means} ($k=20$) \\
        & \texttt{nsamples} & \texttt{auto} \\
        & \texttt{l1\_reg} & \texttt{auto} \\
        \bottomrule
    \end{tabular}
    \caption{Explainer hyperparameters used across all experiments. In general, the default explainer hyperparameters are used.
    See the documentation of \LIME{} and \SHAP{} for the meaning of each hyperparameter\protect\footnotemark{}.
    }
    \label{tab:explainer_hparams}
\end{table}
\footnotetext{\LIME{}: \href{https://lime-ml.readthedocs.io/en/latest/}{https://lime-ml.readthedocs.io/en/latest/}; \SHAP{}: \href{https://shap.readthedocs.io/en/latest/index.html}{https://shap.readthedocs.io/en/latest/index.html}}

\begin{table}[H]
    \small
    \centering
    \begin{tabular}{@{}lcr@{}}
        \toprule
        Algorithm & Hyperparameter & Value \\
        \midrule
        \multirow{4}{*}{\KNNCAD{}}
        & $k$ & 15 \\
        & $\phi$ & \texttt{max} \\
        & $\epsilon$ & 0.1 \\
        & \texttt{distance} & \texttt{minkowski} \\
        & $p$ & 1 \\
        \midrule
        \multirow{3}{*}{\DETECT{}}
        & $\tau_{\text{global}}$ & $[0.11,0.12]$ \\
        & $n_{\text{train}}$ & $0.9 \times |\bm{\mathcal{D}}|$ \\
        & $n_p$ & $\min(|\bm{\mathcal{X}}^{(g)}|, |\bm{\mathcal{D}}_{\text{train}}| \times 10)$ \\
        \midrule
        \multirow{1}{*}{\DEFENSE{}}
        & $\tau$ & 0.75 \\
        \bottomrule
    \end{tabular}
    \caption{Defense algorithm hyperparameters used for each experiment unless otherwise specified.
    }
    \label{tab:defense_hparams}
\end{table}

\paragraph{Hardware and Software}
All experiments were run on a machine with the following relevant hardware and software specifications:

\begin{itemize}
    \item Hardware
    \begin{itemize}
        \item AMD Ryzen 9 5950X (code is CPU only)
        \item 64 GiB DDR4 RAM
    \end{itemize}
    \item Software
    \begin{itemize}
        \item Manjaro Linux 21.2.6 (Qonos)
        \item Python 3.9.13
        \begin{itemize}
            \item ipykernel==6.7.0
            \item jupyter-client==7.1.2
            \item jupyter-core==4.9.1
            \item lime==0.2.0.1
            \item matplotlib==3.5.1
            \item numpy==1.21.0
            \item pandas==1.4.0
            \item scikit-learn==1.0.2
            \item scipy==1.7.3
            \item seaborn==0.11.2
            \item shap==0.40.0
        \end{itemize}
    \end{itemize}
\end{itemize}

\section{Data}

\paragraph{COMPAS}
The COMPAS data set is available for download on the \textit{ProPublica} GitHub repository. The data set is provided by \textit{ProPublica} without a listed license but external analysis of the data is encouraged. The data set does not contain offensive content and does not contain personally identifiable information --- all data is collected from public records of the defendants in Broward County, Florida.

\href{https://github.com/propublica/compas-analysis}{https://github.com/propublica/compas-analysis}

\paragraph{German Credit}
The German Credit data set is available through the UCI machine learning repository. UCI-donated data sets are licensed under the Creative Commons Attribution 4.0 International license (CC BY 4.0). UCI-donated data sets also do not have personally-identifiable information. The data set does not contain offensive content.

\href{https://archive.ics.uci.edu/ml/datasets/statlog+(german+credit+data)}{https://archive.ics.uci.edu/ml/datasets/statlog+(german+credit+data)}

\paragraph{Communities \& Crime}
The Communities \& Crime data set is available through the UCI machine learning repository. UCI-donated data sets are licensed under the Creative Commons Attribution 4.0 International license (CC BY 4.0). UCI-donated data sets also do not have personally-identifiable information. The data set does not contain offensive content.

\href{https://archive.ics.uci.edu/ml/datasets/communities+and+crime}{https://archive.ics.uci.edu/ml/datasets/communities+and+crime}

\section{Error Bars}\label{sec:error_bars}
While our core algorithm, \KNNCAD{}, is deterministic, there is randomness in experiments owing to the perturbation processes of \LIME{} and \SHAP{}.
$f_{\text{biased}}$ is deterministic as it is rule-based and operates on feature values that do not change between trials.
While $f_{\text{unbiased}}$ is deterministic, its predictions are nondeterministic due to the random generation process of the uncorrelated features used in the COMPAS and Communities \& Crime tasks.
Furthermore, $f$ is not deterministic as $\dood{}$ is a function of a random forest.
In turn, both \DETECT{} and \DEFENSE{} are nondeterministic.
We report the results with standard deviations across 10 trials in Table~\ref{tab:results_error_bars}.

\input{tables/results_error_bars}

\section{Additional Top-3 Feature Plots}

\begin{figure}[H]
    \centering
    \includegraphics[width=\linewidth]{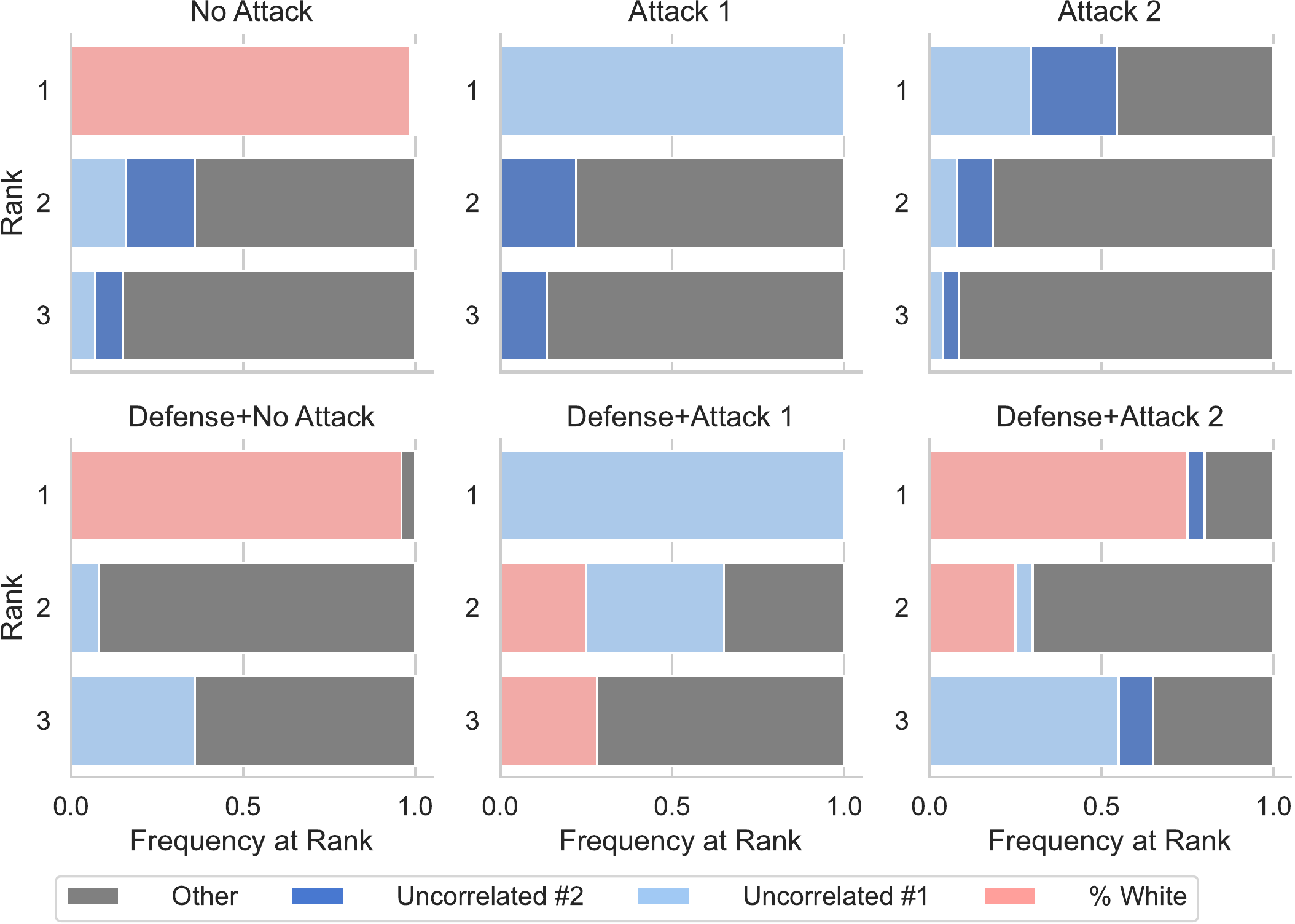}
    \caption{The top-3 most important features identified by \LIME{} with and without the defense on the Communities \& Crime task. Our defense improves upon the first attack and mitigates the second attack and demonstrates that \ACME{} likely attempted to conceal that race is frequently factored into violent crime prediction.}
    \label{fig:topk-lime-cc}
\end{figure}

\begin{figure}[H]
    \centering
    \includegraphics[width=\linewidth]{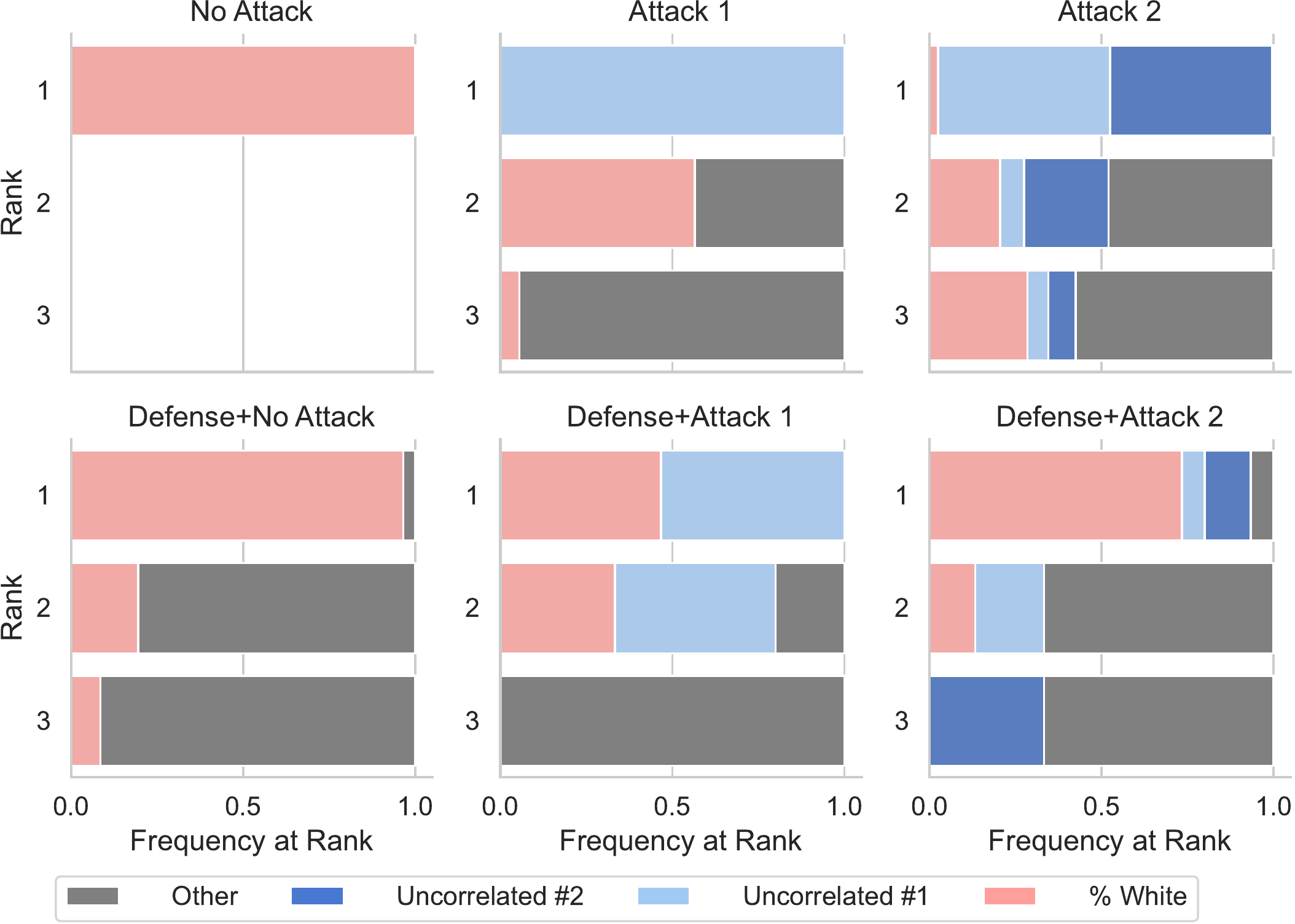}
    \caption{The top-3 most important features identified by \SHAP{} with and without the defense on the Communities \& Crime task. Our defense improves upon the first attack and mitigates the second attack and demonstrates that \ACME{} likely attempted to conceal that race is frequently factored into violent crime prediction.}
    \label{fig:topk-shap-cc}
\end{figure}

\begin{figure}[H]
    \centering
    \includegraphics[width=\linewidth]{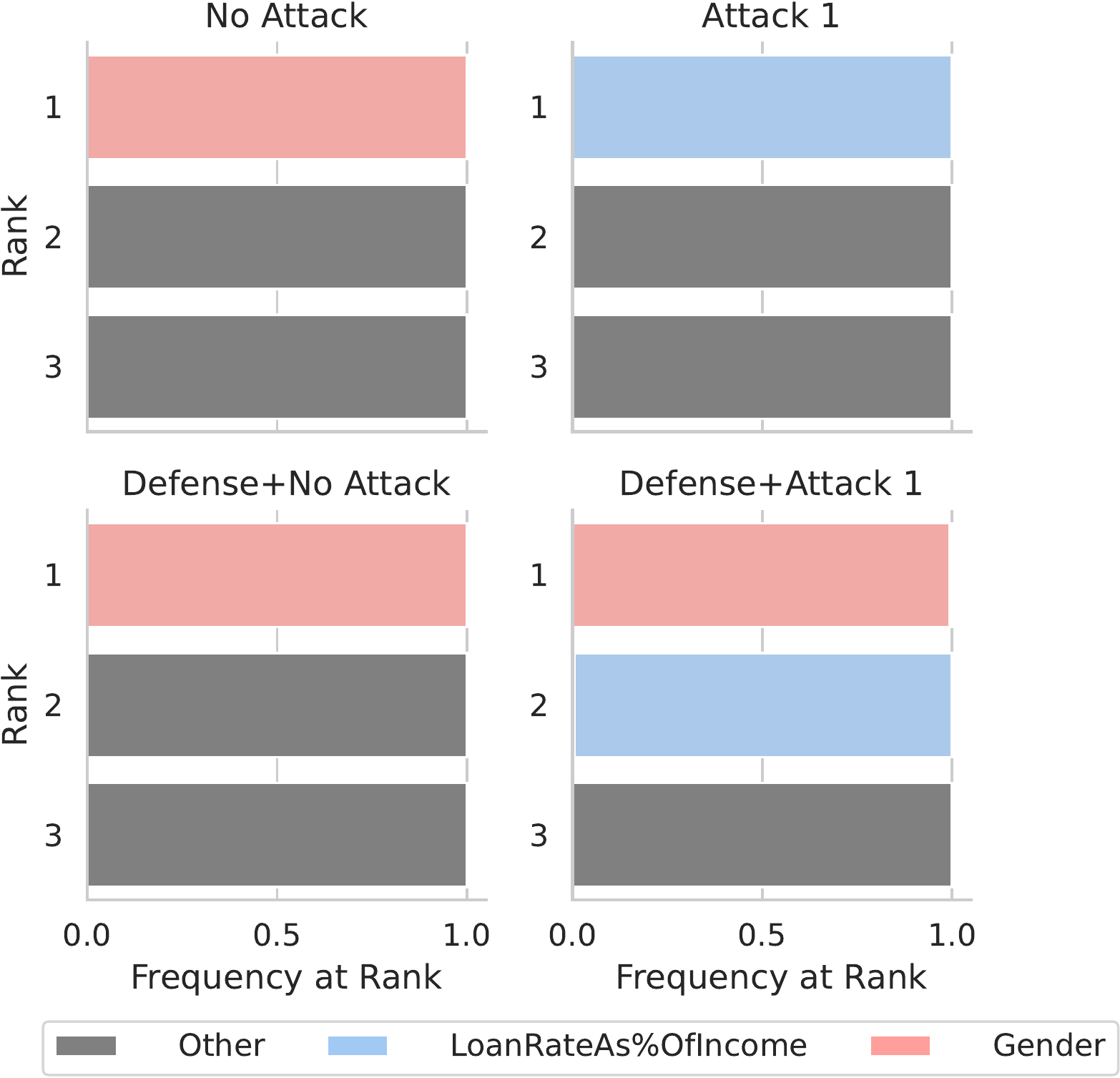}
    \caption{The top-3 most important features identified by \LIME{} with and without the defense on the Boston Credit task. Our defense mitigates the attack and demonstrates that \ACME{} attempted to conceal that gender is frequently factored into credit risk.}
    \label{fig:topk-lime-boston}
\end{figure}

\begin{figure}[H]
    \centering
    \includegraphics[width=\linewidth]{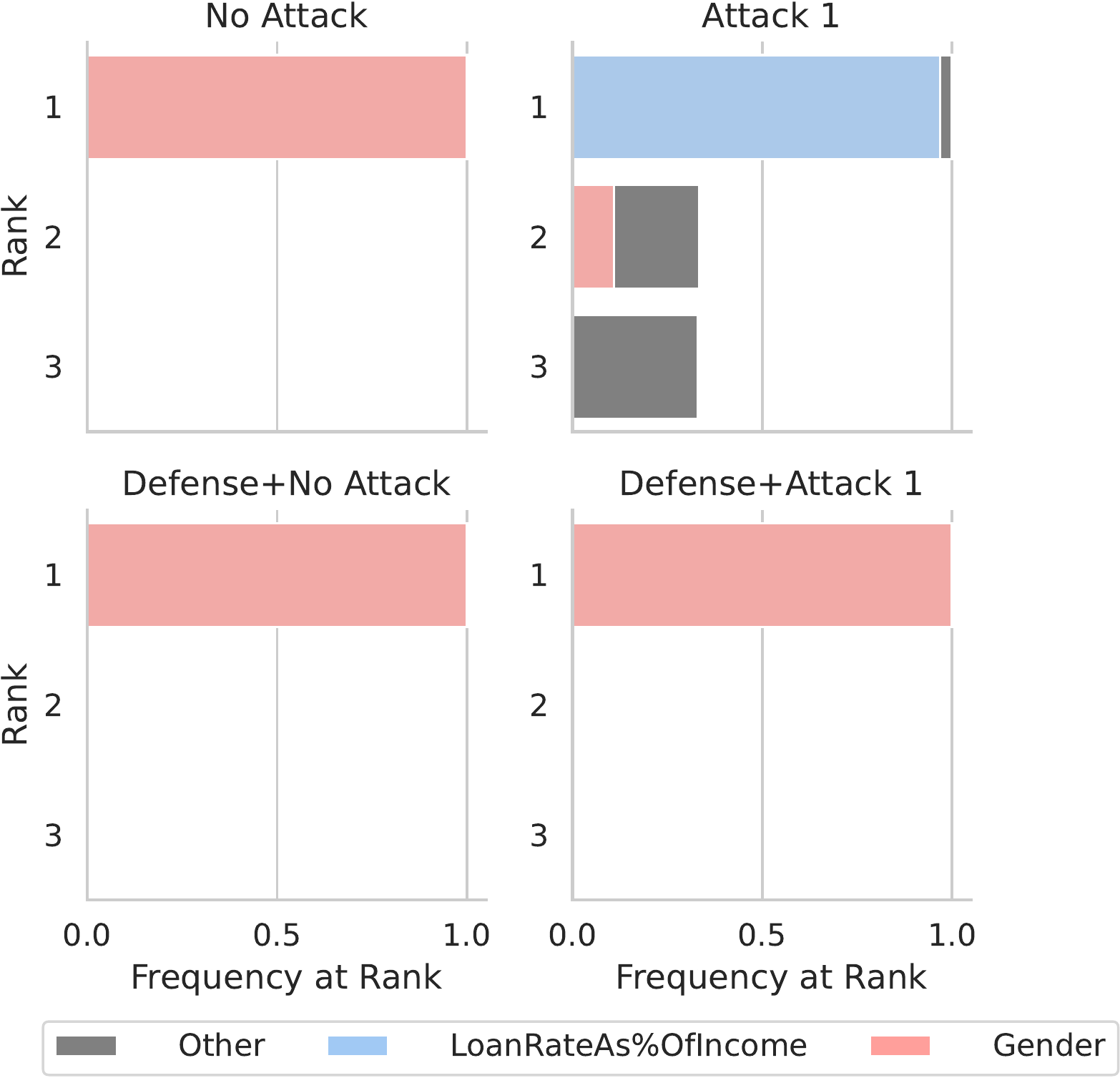}
    \caption{The top-3 most important features identified by \SHAP{} with and without the defense on the Boston Credit task. Our defense mitigates the attack and demonstrates that \ACME{} attempted to conceal that gender is frequently factored into credit risk.}
    \label{fig:topk-shap-boston}
\end{figure}

The following figures show the top-3 most important features as identified by \LIME{} and \SHAP{} with and without the defense across all attack scenarios: Figures~\ref{fig:topk-lime-cc}-\ref{fig:topk-shap-boston}.
For the Communities and Crime attack with \LIME{} and $N_{hl}=1$, our method is able to increase the importance of the sensitive feature in the second rank, but not the first. However, \DETECT{} is still able to detect that $f$ is acting adversarially as shown in Table~\ref{tab:results_error_bars}. Otherwise, our approach is successful.

\section{Additional Analyses}

\begin{figure}[H]
    \centering
    \includegraphics[width=\linewidth]{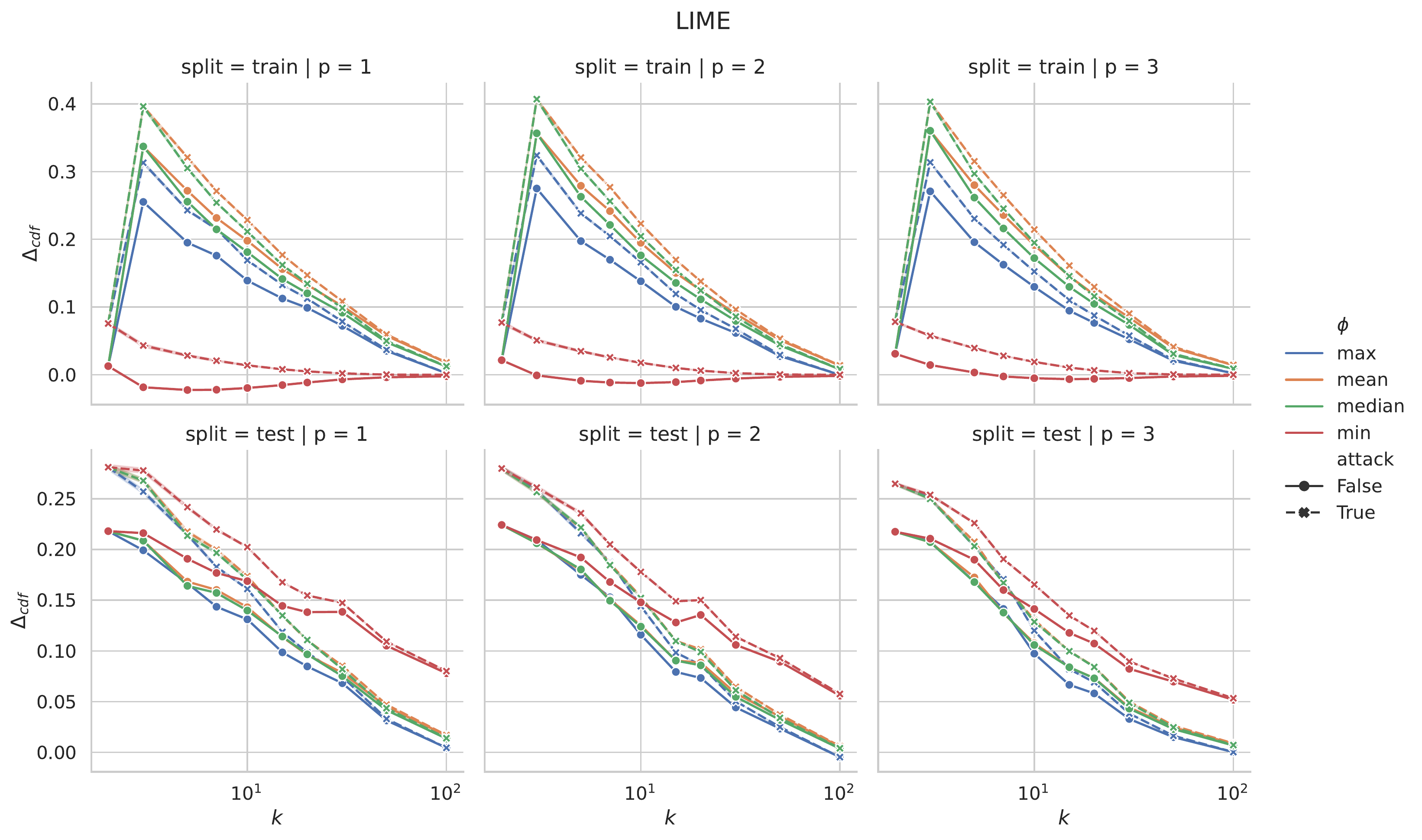}
    \caption{Analysis of hyperparameters of \KNNCAD{} when performing detection (\DETECT{}) with the \LIME{} explainer. The $\Delta_{cdf}$ score is reported as a function of $k$ on the Communities \& Crime dataset. The ideal hyperparameters maximize the difference between $\Delta_{cdf}$ when the attack is deployed and is not.}
    \label{fig:hparam_lime}
\end{figure}

\begin{figure}[H]
    \centering
    \includegraphics[width=\linewidth]{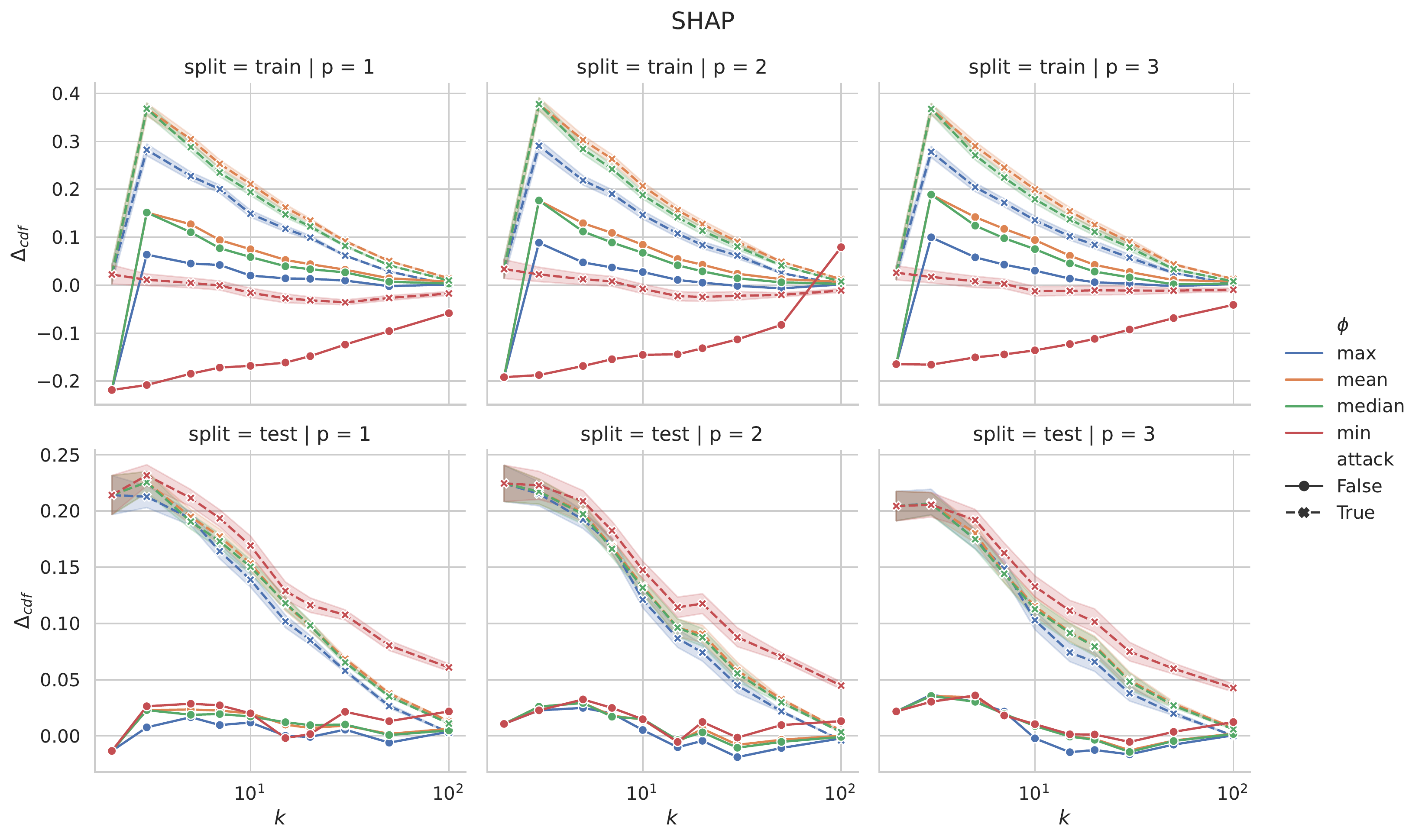}
    \caption{Analysis of hyperparameters of \KNNCAD{} when performing detection (\DETECT{}) with the \SHAP{} explainer. The $\Delta_{cdf}$ score is reported as a function of $k$ on the Communities \& Crime dataset. The ideal hyperparameters maximize the difference between $\Delta_{cdf}$ when the attack is deployed and is not.}
    \label{fig:hparam_shap}
\end{figure}

\begin{figure}[H]
    \centering
    \includegraphics[width=\linewidth]{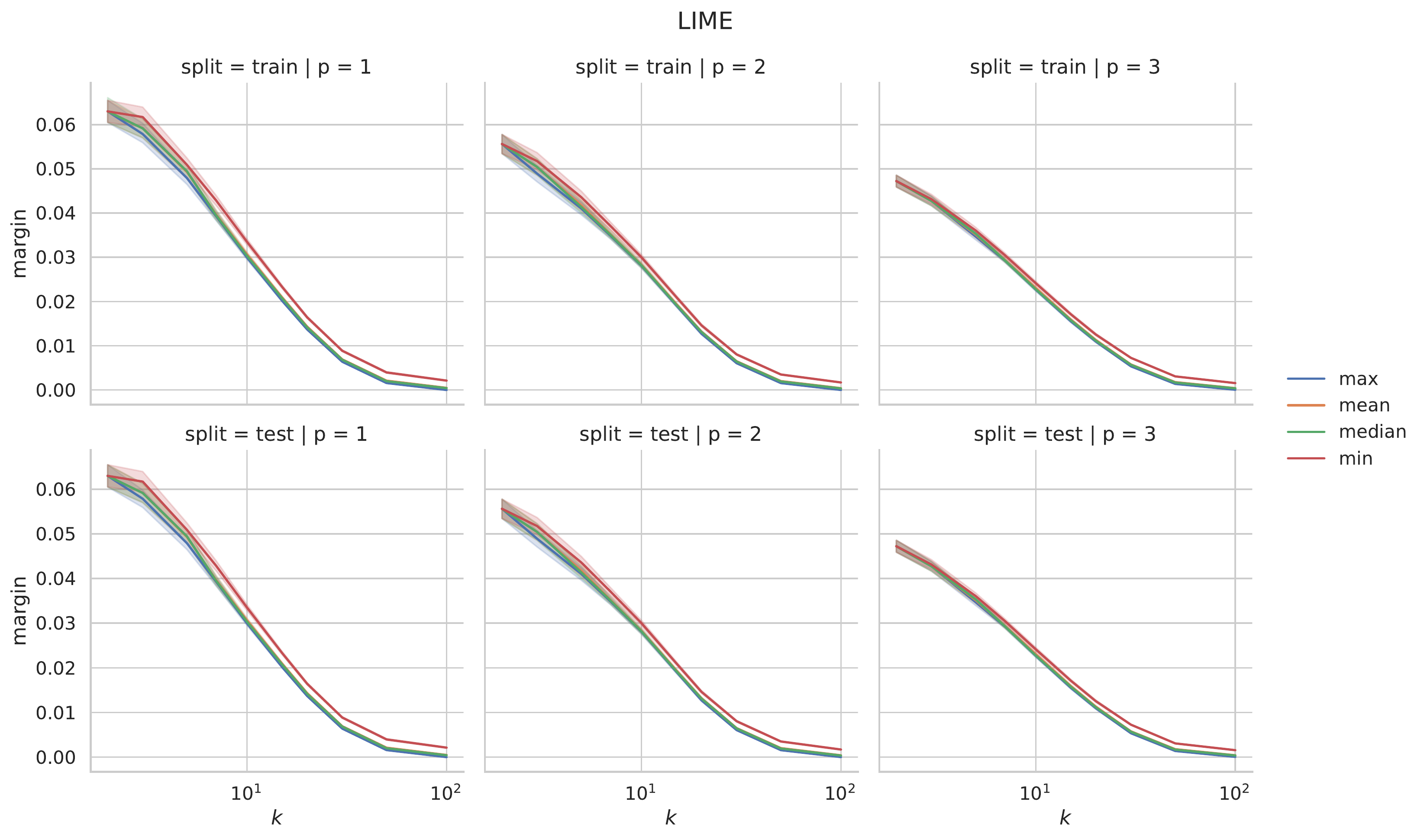}
    \caption{Analysis of hyperparameters of \KNNCAD{} when performing detection (\DETECT{}) with the \LIME{} explainer. The margin is reported as a function of $k$ on the Communities \& Crime dataset.}
    \label{fig:margin_hparam_lime}
\end{figure}

\begin{figure}[H]
    \centering
    \includegraphics[width=\linewidth]{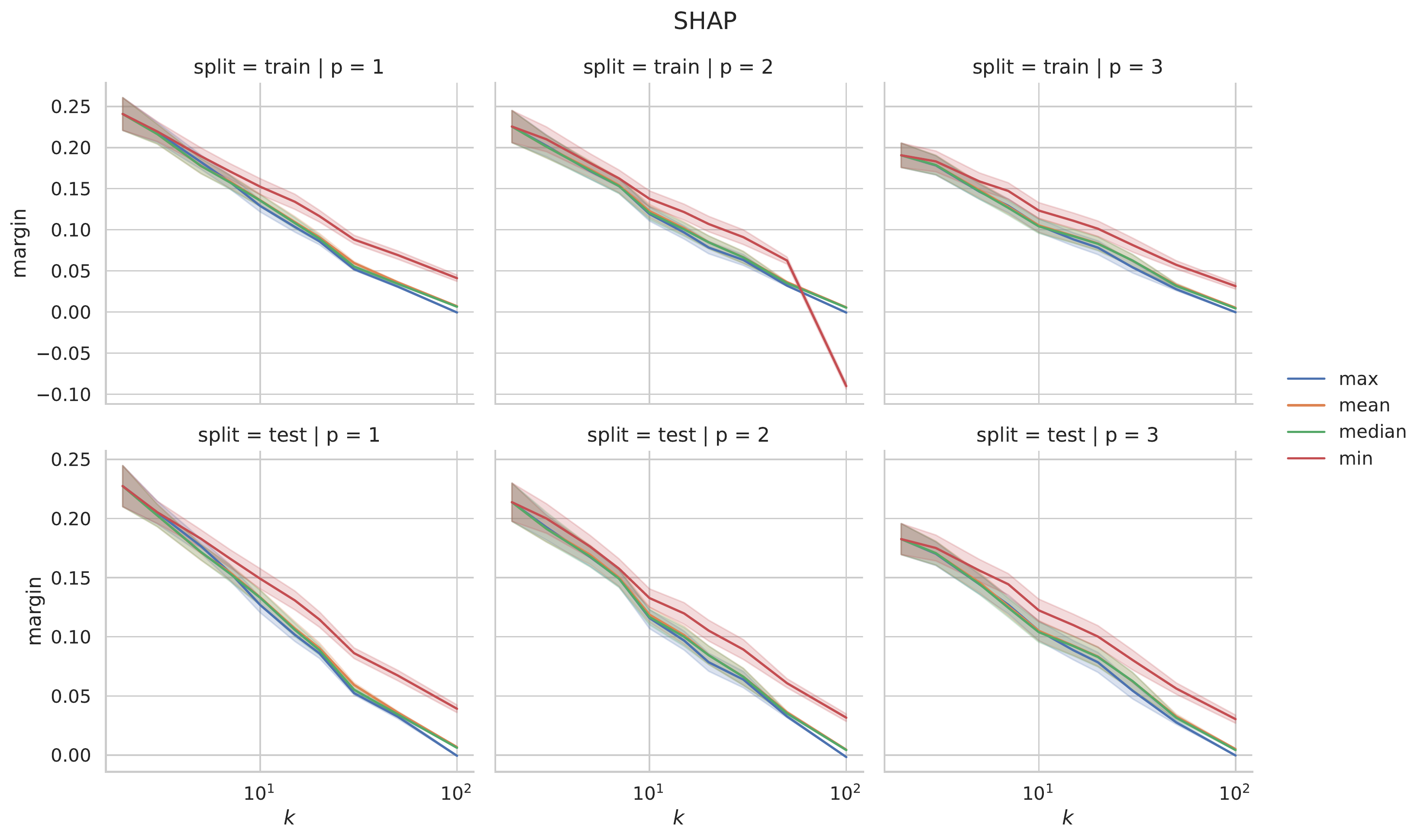}
    \caption{Analysis of hyperparameters of \KNNCAD{} when performing detection (\DETECT{}) with the \SHAP{} explainer. The margin is reported as a function of $k$ on the Communities \& Crime dataset.}
    \label{fig:margin_hparam_shap}
\end{figure}

\paragraph{Hyperparameter Analysis}
Here, we conduct an analysis of the \KNNCAD{} hyperparameters $\phi$, $k$, and Minkowski distance $p$ on the Communities \& Crime data set ---
see Figures~\ref{fig:hparam_lime} and~\ref{fig:hparam_shap}.
Ideally, \KNNCAD{} maximizes the difference between $\Delta_{cdf}$ when the attack is deployed and $\Delta_{cdf}$ when no attack is present. We refer to this difference as the \textit{margin}.
The margin is visualized as a function of the same hyperparameters in Figures~\ref{fig:margin_hparam_lime} and~\ref{fig:margin_hparam_shap}.
The $\Delta_{cdf}$ scores in both cases decrease as $k$ increases. Irrespective of $\phi$, $\Delta_{cdf}$ is the same when $k=1$. However, $\Delta_{cdf}$ scores peak when $k>1$.
There is very little difference in $\Delta_{cdf}$ or the margin for different values of $p$ (although $p=1$ results in a slightly higher peak margin). The margin is maximized with lower values of $k$. The selection of $\phi$ generally has the affect of modulating the mean value of $\Delta_{cdf}$. With the train split, $\phi=\texttt{min}$ results in a abnormally low mean $\Delta_{cdf}$ simply because the neighbors are of the same split. All values of $\phi$ result in a similar margin, but $\phi = \texttt{min}$ performs best across all values of $k$, $p$, and split. An oddity is that $\phi = \texttt{min}$ begins to fail for \SHAP{} on the train split with $p=2$ and $k > 50$ -- we presume this occurs because $\phi = \texttt{min}$ is a brittle estimator compared to the mean or median.
All of these observations hold between both the train and test splits used to estimate $\Delta_{cdf}$ in the \DETECT{} algorithm.
Owing to its more principled perturbation scheme, the margin when using \SHAP{} is ${\sim}5{\times}$ greater than that of \LIME{} in general.

\section{Further Delineating the Problem Scenario Formalism}

In this appendix, we elucidate our formalization of the attack and defense scenarios using a rolling example. Please refer back to Figure 1 in the main text for a visual aid of the scenario.

The potential adversary, \ACME{} Corporation, has an automated decision-making system $f$. The system $f$ makes decisions on behalf of its users in potentially high-stakes applications and is a black box. The data (denoted by $\bm{\mathcal{X}}$) comprises $F$ features, some of which may be protected (or sensitive) according to regulation or law. To ensure that each algorithm $f$ deployed by \ACME{} meets regulatory compliance, an auditor devises a strategy to understand the decision-making process of $f$.

In this work, it is considered that $f$ fails to meet compliance if it uses a protected feature to make high-stakes decisions, and if $f$ behaves adversarially depending on whether it is being audited. To verify that \ACME{} meets such compliance, the auditor uses an explainer, $g$, that provides a set of feature contributions $\{a_{ij}\}_j^F$ for the $i^{\text{th}}$ data sample $\mathbf{x}_i$. Each contribution $a_{ij}$ provides the additive contribution of feature $j$ to the prediction value of $f(\mathbf{x}_i)$. In the main text, the explainer $g$ produces explanations by optimizing the objective defined in Eq.\ (1). Since the explainers are perturbation-based, a set of perturbed data samples are generated when explaining data. We denote this set of perturbations as $\bm{\mathcal{X}}^{(g)}$ and note that it is derived from $\bm{\mathcal{X}}$. The auditor thereafter can determine if \ACME{} met compliance if the features were not used in a way that violates regulations.

Unfortunately, \ACME{} may act in an adversarial manner and attempt to hide the decision-making process of its algorithm $f$. For instance, a scaffolding attack may be employed as described in Section~2 and Eq.\ (2) -- if \ACME{} detects that an explainer $g$ is querying $f$, it can toggle from its biased model to one that is unbiased, thwarting the audit. The general objective of an attack is laid out in Eq.\ (4). In the equation, recall that $\mathcal{E}_i^{-} = \mathcal{E}_i^{\text{sensitive}} \cup \mathcal{E}_i^{\text{harmless}} = g(\mathbf{x}_i; f)$ is an ordered set of feature contributions for the $i^\text{th}$ data sample, and $\mathcal{E}_i^{\text{sensitive}}$ and $\mathcal{E}_i^{\text{harmless}}$ are the subsets of $\mathcal{E}_i^{-}$ corresponding to the sensitive and harmless features, respectively. With that established, the objective is to simultaneously minimize the error between the ground truth, $y_i$, and the predictions of $f$ (as is done in normal supervised learning), but to also minimize the number of highly-ranked contributions of sensitive features. Put even simpler, an adversarial \ACME{} wants to retain the fidelity of its algorithm $f$ while hiding the fact that sensitive features are important to the decision-making process.

In this work, we consider that an adversarial \ACME{} employs the scaffolding attack. The objective of the attack is defined in Eq.\ (5). Let us assume that we, as an oracle, know whether a data sample originated from normal data ($\bm{\mathcal{X}}$) or from the perturbations of an explainer $g$ ($\bm{\mathcal{X}}^{(g)}$). Let $t_i \in \mathcal{T}$ denote this information, i.e., $t_i = 1$ if $\mathbf{x}_i \in \bm{\mathcal{X}}^{(g)}$ and $t_i = 0$ otherwise. With this established, \ACME{} desires to use a biased algorithm $f_{\text{biased}}$ whenever $t_i = 0$ and an unbiased algorithm $f_{\text{unbiased}}$ when $t_i = 0$. In the scaffolding approach, this is solved by selecting the out-of-distribution detector, $\dood{}$, that minimizes Eq.\ (5). In short, the adversary aims to use an unbiased predictor when it is audited in order to meet regulatory compliance.

As discussed, the auditor can use $g$ to provide explanations of the decisions of $f$ to see if the way features were used violates regulatory compliance. However, we did not establish how the auditor can verify that $f$ does not change its decision-making process when being audited. Without a defense approach, \ACME{} can hide their decision-making processes from the auditor as-is. We provide an approach to the detection and defense of the scaffolding attack, the former detecting whether the attack is deployed and the latter mitigating the attack so that explanations are not altered. The problem can be understood by considering the conditional probability mass function $p(f(\mathbf{x}_i) \mid \mathbf{x}_i)$. The distribution indicates the likelihood of $f$ predicting some outcome given that the input is $\mathbf{x}_i$, which is necessary because the attacker changes its prediction (e.g., biased or unbiased) based on the input (e.g., on- or off-manifold). Based on the probability measure of individual samples, i.e., $\mathbb{P}(f(\mathbf{x}_i) \mid \mathbf{x}_i)$, we can determine if the behavior of $f$ has changed -- low-likelihood samples indicate that $f$ is likely behaving adversarially. In the defense approach, $p$ is estimated by a conditional anomaly detector $h$ that should optimize for the objective given by Eq.\ (6). Put simply, the goal of the defense is to learn the distribution of normal behavior of $f$. If $f$ is not acting normally (the conditional likelihood of predictions is below a threshold as in Eq.\ (3)), then we can perform both detection and explanation defense. Detection is performed as in Eq.\ (7) where we measure the distance between the distributions $p( f(\mathbf{x}_i) \mid \mathbf{x}_i )$ and $p( f(\mathbf{x}^{(g)}_i) \mid \mathbf{x}^{(g)}_i )$ -- if the distribution of perturbed samples differs significantly from natural samples, then $f$ is likely acting adversarially for the explainer $g$. At this point, the auditor can determine whether \ACME{} is complying with the audit and whether to trust the explanations of $g$ before further scrutiny.

For descriptions of the \KNNCAD{} and explanation defense approaches, please see ``Detection, Defense and KNN-CAD'' and Algorithms 1--4 in Section 3 of the main text.

\section{Additional Attack Vectors}

It is possible that the attacker can detect whether multiple queries come from the same neighborhood consecutively. If so, the attacker can then exploit this information to always use the unbiased model. We can easily devise a solution to mitigate this attack that follows the defense approach. Simply, the queries that are made to $f$ can be reordered to alternate between neighborhoods. Since there are multiple samples that are being explained, there is in turn a neighborhood for each sample. By switching between multiple neighborhoods, this makes detecting neighborhoods a substantially more difficult problem. Naturally, since the number of explainer perturbations outnumber the samples representing normality, one might wonder if there is a tractable number of samples in order for the defense to be undetected in this attack vector. In Appendix~F, we demonstrate that our approach is extremely sample-efficient -- a small percentage of the neighborhood samples are required to be confident in whether $f$ is behaving adversarially. To realize the defense against this attack vector, we modify Algorithms~3 and 4 to perform a shuffle stratified among the neighborhoods.

\section{Time and Space Complexities}

\paragraph{Algorithm \KNNCAD{}{\small\texttt{.fit}}}\mbox{}\\

\textbf{Time}: $\mathcal{O}(F N \log N)$ (construction of a ball tree) + $\mathcal{O}(N (k \log N + T_{f}(F)))$ (\KNNCAD{}{\small\texttt{.score\_samples}}) + $\mathcal{O}(N \log N)$ (sorting the scores) = $\mathcal{O}(N (F \log N + k \log N + T_{f}(F)))$.

\textbf{Space}: $\mathcal{O}(F N)$ (construction of a ball tree) + $\mathcal{O}(N (k + S_{f}(F)))$ (\KNNCAD{}{\small\texttt{.score\_samples}}) + $\mathcal{O}(N)$ (the scores) = $\mathcal{O}(N (F + k + S_{f}(F)))$.

\paragraph{Algorithm \KNNCAD{}{\small\texttt{.score\_samples}}}\mbox{}\\

\textbf{Time}: $\mathcal{O}(N \,T_{f}(F))$ (calling $f$ for each sample) + $\mathcal{O}(k N \log N)$ (querying the $k$ nearest neighbors for each sample) + $\mathcal{O}(N k)$ (computing scores from neighbors and classes for each sample) = $\mathcal{O}(N (k \log N + T_{f}(F)))$. In practice where $T_{f}(F) \gg k \log N$ is likely, \eg{}, with DNNs, the time complexity is $\mathcal{O}(N T_{f}(F))$. If we are provided with pre-computed predictions, the time complexity is $\mathcal{O}(k N \log N)$.

\textbf{Space}: $\mathcal{O}(N \,S_{f}(F))$ (calling $f$ for each sample) + $\mathcal{O}(k N)$ (querying the $k$ nearest neighbors for each sample) + $\mathcal{O}(N)$ (computing scores from neighbors and classes for each sample) = $\mathcal{O}(N (k + S_{f}(F)))$. In practice where $S_{f}(F) \gg k$ is likely, the space complexity is $\mathcal{O}(N S_{f}(F))$. If we are provided with pre-computed predictions, the space complexity is $\mathcal{O}(k N)$.

\paragraph{Algorithm \DETECT{}}\mbox{}\\

\textbf{Time}:
$\mathcal{O}(N (F \log N + k \log N + T_{f}(F)))$ (fitting \KNNCAD{}) +
$\mathcal{O}(N (k \log N + T_{f}(F)))$ (scoring the test samples) +
$\mathcal{O}(N n_p (k \log (N n_p) + T_{f}(F)))$ (scoring the $n_p$ perturbations per test sample) +
$\mathcal{O}(N n_p)$ (computing areas)
= $\mathcal{O}(N n_p (k \log (N n_p) + T_{f}(F)))$.

\textbf{Space}:
$\mathcal{O}(N (F + k + S_{f}(F)))$ (fitting \KNNCAD{}) +
$\mathcal{O}(N (k + S_{f}(F)))$ (scoring the test samples) +
$\mathcal{O}(N n_p (k + S_{f}(F)))$ (scoring the $n_p$ perturbations per test sample) +
$\mathcal{O}(N n_p)$ (computing areas)
= $\mathcal{O}(N n_p (k + S_{f}(F)))$.

\paragraph{Algorithm \DEFENSE{}}\mbox{}\\

\textbf{Time}:
$\mathcal{O}(n_p (k \log n_p + T_{f}(F)))$ (scoring the $n_p$ perturbations for the test sample) +
$\mathcal{O}(n_p)$ (filtering and set operations) +
$\mathcal{O}(R \, T_{\DEFENSE{}}(n_p, k, F))$ (the algorithm is executed in $R$ recursions, with the same input size in the worst case)
= $\mathcal{O}(R n_p (k \log n_p + T_{f}(F)))$. In the worst case, all generated samples are abnormal and filtered, \eg{}, if the threshold is set too high or $h$ is fit poorly. In practice, fewer than $n_p$ samples are required at each recursion.

\textbf{Space}:
$\mathcal{O}(n_p (k + S_{f}(F)))$ (scoring the $n_p$ perturbations for the test sample) +
$\mathcal{O}(n_p)$ (filtering and set operations)
= $\mathcal{O}(n_p (k + S_{f}(F)))$. The space of only one recursion is necessary to consider.

%% file: tables/results_error_bars.tex
\begin{table*}%
    \small
    \centering
    \ra{1.2}
    \begin{tabular}{@{}lcccccccccc@{}}
        \toprule
        \multirow{2}{*}{$g$} & \multirow{2}{*}{\rotatebox{90}{Data Set}} & \multirow{2}{*}{$N_{hl}$}
        &\multicolumn{4}{c}{Fidelity}
        & {\DETECTtiny{}} & \multicolumn{2}{c}{\hspace{1.75ex}\DEFENSEtiny{}} \\
        \cmidrule(lr){4-7}\cmidrule(lr){8-8}\cmidrule(lr){9-10}
        &&& ${f}$ & ${\dood{}}$ & ${g}$ & ${h}$ & $\Delta_{\text{cdf}}$ & $\text{inf}_{g}$ & $\text{fid}_{f}$ \\
        \midrule
        \multirow{8}{*}{\rotatebox{90}{\LIME{}}}&\multirow{3}{*}{CO}%
        & --
        & 1.00${\pm}$0.00 & --   & 0.31${\pm}$0.01 & 0.93${\pm}$0.01 & 0.11${\pm}$0.01 & 0.00${\pm}$0.00 & 0.31${\pm}$0.01 \\
        && 1
        & 0.99${\pm}$0.01 & 0.99${\pm}$0.01 & 0.23${\pm}$0.01 & 0.81${\pm}$0.01 & 0.30${\pm}$0.02 & 0.05${\pm}$0.01 & 0.30${\pm}$0.01\\
        && 2
        & 0.99${\pm}$0.01 & 0.99${\pm}$0.01 & 0.10${\pm}$0.01 & 0.80${\pm}$0.01 & 0.30${\pm}$0.01 & 0.00${\pm}$0.00 & 0.31${\pm}$0.01\\[1.25ex]
        &\multirow{2}{*}{GC}%
        & --
        & 1.00${\pm}$0.00 & --   & 0.27${\pm}$0.01 & 0.78${\pm}$0.01 & 0.07${\pm}$0.02 & 0.00${\pm}$0.00 & 0.27${\pm}$0.01\\
        && 1
        & 1.00${\pm}$0.00 & 0.99${\pm}$0.01 & 0.18${\pm}$0.01 & 0.70${\pm}$0.02 & 0.18${\pm}$0.02 & 0.00${\pm}$0.00 & 0.27${\pm}$0.01\\[1.25ex]
        &\multirow{3}{*}{CC}%
        & --
        & 1.00${\pm}$0.00 & --   & 0.21${\pm}$0.01 & 0.80${\pm}$0.01 & 0.09${\pm}$0.01 & 0.00${\pm}$0.00 & 0.16${\pm}$0.01\\
        && 1
        & 1.00${\pm}$0.00 & 0.99${\pm}$0.01 & 0.16${\pm}$0.01 & 0.79${\pm}$0.02 & 0.12${\pm}$0.01 & 0.01${\pm}$0.00 & 0.20${\pm}$0.01\\
        && 2
        & 1.00${\pm}$0.00 & 0.99${\pm}$0.01 & 0.17${\pm}$0.02 & 0.79${\pm}$0.03 & 0.12${\pm}$0.01 & 0.00${\pm}$0.00 & 0.31${\pm}$0.01\\
        \midrule
        \multirow{8}{*}{\rotatebox{90}{\SHAP{}}}&\multirow{3}{*}{CO}%
        & --
        & 1.00${\pm}$0.00 & --   & 0.29${\pm}$0.01 & 0.99${\pm}$0.01 & -0.02${\pm}$0.01 & 0.00${\pm}$0.00 & 0.29${\pm}$0.01 \\
        && 1
        & 0.93${\pm}$0.01 & 0.82${\pm}$0.02 & 0.24${\pm}$0.01 & 0.77${\pm}$0.02 & 0.20${\pm}$0.01 & 0.02${\pm}$0.00 & 0.28${\pm}$0.01 \\
        && 2
        & 0.91${\pm}$0.01 & 0.82${\pm}$0.02 & 0.24${\pm}$0.01 & 0.77${\pm}$0.02 & 0.25${\pm}$0.01 & 0.01${\pm}$0.00 & 0.26${\pm}$0.02 \\[1.25ex]
        &\multirow{2}{*}{GC}%
        & --
        & 1.00${\pm}$0.00 & --   & 0.24${\pm}$0.01 & 0.86${\pm}$0.01 & -0.05${\pm}$0.00 & 0.00${\pm}$0.00 & 0.27${\pm}$0.02 \\
        && 1
        & 0.83${\pm}$0.02 & 0.71${\pm}$0.02 & 0.12${\pm}$0.01 & 0.61${\pm}$0.02 & 0.12${\pm}$0.01 & 0.00${\pm}$0.00 & 0.27${\pm}$0.02 \\[1.25ex]
        &\multirow{3}{*}{CC}%
        & --
        & 1.00${\pm}$0.00 & --   & 0.08${\pm}$0.01 & 0.85${\pm}$0.01 & 0.00${\pm}$0.01 & 0.00${\pm}$0.00 & 0.20${\pm}$0.01 \\
        && 1
        & 0.97${\pm}$0.01 & 0.85${\pm}$0.02 & 0.07${\pm}$0.00 & 0.79${\pm}$0.01 & 0.13${\pm}$0.01 & 0.01${\pm}$0.00 & 0.19${\pm}$0.01 \\
        && 2
        & 0.99${\pm}$0.01 & 0.85${\pm}$0.02 & 0.07${\pm}$0.00 & 0.77${\pm}$0.02 & 0.12${\pm}$0.01 & 0.00${\pm}$0.00 & 0.30${\pm}$0.01 \\
        \bottomrule
    \end{tabular}
    \caption{The (in)fidelity and $\Delta_{\text{cdf}}$ scores of the attack and defense across the real-world data sets when \KNNCAD{} is used as $h$. The scores are reported across 10 trials with standard deviations.
    The data sets are abbreviated as follows: COMPAS (CO), German Credit (GC), Communities \& Crime (CC).
    $N_{hl}$ is the number of harmless features used by $f_{\text{unbiased}}$ and is unspecified when no attack is deployed. See the text for the differing definitions of fidelity for each algorithm. We abbreviate $\text{infidelity}_{\DEFENSE{},g}$ as $\text{inf}_{g}$ and $\text{fidelity}_{\DEFENSE{},f}$ as $\text{fid}_{f}$. Our approach successfully detects the scaffolding attack while maintaining explanation fidelity for both explainers.}
    \label{tab:results_error_bars}
\end{table*}